\documentclass[journal]{IEEEtranTIE}
\usepackage{graphicx}
\usepackage{cite}
\usepackage{picinpar}
\usepackage{amsmath}
\usepackage{url}
\usepackage{flushend}
\usepackage[latin1]{inputenc}
\usepackage{colortbl}
\usepackage{soul}
\usepackage{multirow}
\usepackage{pifont}
\usepackage{color}
\usepackage{alltt}
\usepackage[hidelinks]{hyperref}
\usepackage{enumerate}
\usepackage{siunitx}
\usepackage{epstopdf}
\usepackage{pbox}
\usepackage{amsfonts}
\usepackage{bm}
\usepackage{empheq}
\usepackage{amsthm}
\usepackage[caption=false, font=footnotesize]{subfig}
\usepackage{algorithm2e}
\newtheoremstyle{assumptionstyle}
{3pt}
{3pt}
{\normalfont}
{\parindent}
{\itshape}
{.}
{.5em}
{}
\theoremstyle{assumptionstyle}
\newtheorem{assumption}{Assumption}

\usepackage{mdframed}
\usepackage{xcolor}

\newmdenv[
  backgroundcolor=yellow,
  linecolor=white,
  innertopmargin=0pt,
  innerbottommargin=0pt,
  innerleftmargin=0pt,
  innerrightmargin=0pt,
  skipabove=0pt,
  skipbelow=0pt
]{highlight}

\begin{document}
\title{Human-Robot Interaction via Meta-Learning Based Mode Recognition and Adaptive Control}
\author{
	\vskip 1em
	Nuo~Chen, \emph{Student Member,~IEEE},
	and~Ya-Jun~Pan, \emph{Senior~Member,~IEEE}

	\thanks{
		This work was supported by the Natural Sciences and Engineering Research Council (NSERC) and the Government of Nova Scotia, Canada. (Corresponding author: Ya-Jun Pan)
  
		The authors are with the Advanced Control and Mechatronics Lab, Department of Mechanical Engineering, Dalhousie University, Halifax, B3H 4R2, Canada (e-mails: nuo.chen@dal.ca, yajun.pan@dal.ca).
	}
}
\maketitle
	
\begin{abstract}
In human-robot interaction (HRI) tasks, humans may interact with the robot in either an active guiding or passive following state, resulting in distinct control objectives for each interaction mode. Most existing approaches rely on predefined switching thresholds to distinguish interaction modes, but their performance can be sensitive to measurement noise. To address this limitation, we propose a novel framework with the following key contributions. First, a neural network (NN) trained via meta-learning is developed to recognize human intention without manually defined thresholds. By successfully incorporating force-dependent parameters from the controller, the recognition rates are improved from 90.1\% to 99.1\% for the passive mode and from 79.8\% to 97.1\% for the active mode. Second, an adaptive meta-learning controller is introduced to compensate for human disturbances and enable online adaptation to varying interaction forces. Third, a multiple-trajectory meta-learning algorithm is designed to jointly learn an environment-basis representation and force-dependent parameters, allowing the controller to deal with unmodeled dynamics efficiently. Experimental results demonstrate that the proposed framework achieves robust and accurate control performance, while providing reliable mode recognition for safe and adaptive physical HRI.
\end{abstract}


\markboth{IEEE/ASME TRANSACTIONS ON Mechatronics}%
{}
\definecolor{limegreen}{rgb}{0.2, 0.8, 0.2}
\definecolor{forestgreen}{rgb}{0.13, 0.55, 0.13}
\definecolor{greenhtml}{rgb}{0.0, 0.5, 0.0}

\section{Introduction}
\IEEEPARstart{T}{raditional} robot control and HRI primarily focus on precise motion planning and execution, relying on closed-loop control methods grounded in system dynamics and high-precision sensor feedback. The rapid advancement of technologies such as deep learning and reinforcement learning (RL) has demonstrated remarkable advantages in intelligent control, particularly for systems with high degree-of-freedom (DOF) and in complex environments. However, these approaches often rely heavily on large volumes of high-quality training data. By closely integrating robot control with machine learning (ML), a robust technological foundation can be established to achieve efficient, safe, and flexible HRI across diverse application domains.
\begin{figure*}[t]
    \centering
    \includegraphics[width=0.9\linewidth]{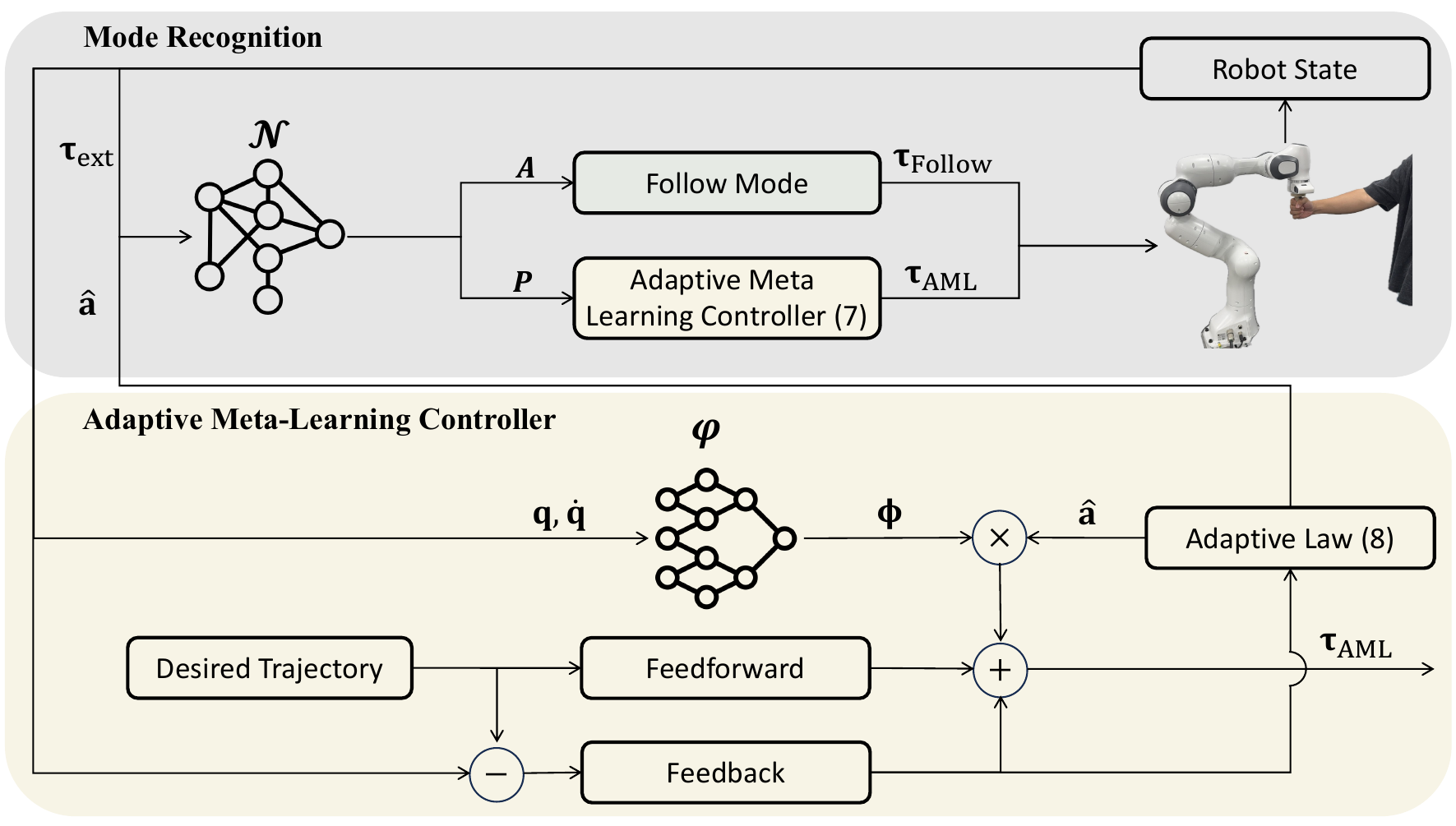}
    \caption{Overview of the proposed framework. The mode recognition NN $\mathcal{N}$ takes the estimated external torque $\tau_{ext}$ and force-dependent parameter $\bm{\hat{a}}$ as inputs to identify human intention during HRI. The robot applies the adaptive meta-learning controller when $\mathcal{N}$ determines that the human is in the passive mode, and switches to the follow mode when $\mathcal{N}$ determines that the human is in the active mode. In the proposed adaptive meta-learning controller, the adaptive law tunes the force-dependent parameter $\hat{\bm{a}}$ to compensate adaptively for unknown external disturbances.}
    \label{fig:Diagram}
\end{figure*}

During physical interaction, robots typically switch between passive and active modes based on system states. Zhou et al. \cite{Zhou_2024_TMech} proposed an active-passive compliance strategy for robotic plugging and unplugging of electrical connectors, integrating a passive compliant device and an active compliance controller with an energy-tank mechanism. This hybrid structure mitigates insertion and withdrawal impacts while maintaining force accuracy. Similarly, Bai et al. \cite{Bai_2021_TRO} developed a spherical wrist with hybrid motion-impedance control, combining mechanical compliance with direction-dependent motion control for unconstrained directions and impedance regulation for constrained ones. In manufacturing, Xu et al. \cite{Xu_2021_RCIM} presented a hybrid active-passive force control strategy for robotic belt grinding of turbine blades, in which active force regulation and passive compliance cooperate to suppress grinding marks. However, these works rely on manually defined thresholds to achieve both high precision and safe physical interaction. Beyond model-based approaches, Bazanella et al. \cite{Bazanella_2025_Arxiv} introduced learning about passivity from data, which is a data-driven framework that identifies system passivity properties directly from input-output measurements via convex optimization, enabling passivity-based stability assurance even for unknown nonlinear systems.

Recent advances in robot control have underscored the growing importance of integrating NN to address challenges associated with nonlinear dynamics, partially known systems, and complex environmental interactions. Yu et al.~\cite{Yu_2024_TIE} proposed an NN-based adaptive human-tracking controller for a flexible articulated machine, achieving high-precision control under uncertain conditions by dynamically adjusting control parameters online. Similarly, NN-based strategies have gained significant attention for trajectory tracking in dynamic environments, as they effectively enhance control accuracy while reducing dependence on precise system models~\cite{Cao_2023_TNNLS}. Hybrid frameworks incorporating sliding mode control (SMC) have also proven effective in mitigating uncertainties during robot-environment interactions, thereby improving adaptability in unstructured environments~\cite{He_2021_TSMC}. These hybrid approaches leverage the complementary strengths of data-driven and model-based techniques. However, key challenges such as improving learning efficiency, ensuring generalization, and enhancing robustness remain critical areas for future research~\cite{Qian_2023_TAI} -\cite{Chen_2024_TIE}. Efficient computation of the dynamics and their derivatives is essential for real-time control. Recent works have proposed analytical approaches for computing derivatives of rigid body dynamics using spatial vector algebra, which benefit for high frequency control and learning based frameworks~\cite{Singh_2022_RAL} -\cite{Singh_2024_TRO}.

Although NNs perform well at modeling unknown dynamics, they require huge amounts of precise data. Such data are difficult to obtain in real-world force-related HRI, as humans cannot simultaneously collect multiple parallel data segments or accelerate the data-gathering process as in simulations. Furthermore, force sensors are expensive, and the acquired data often contain significant noise in both real and simulated environments. Meta-learning has emerged as a promising approach to address the challenge of learning from limited data and rapidly adapting to new tasks. It aims to develop a general learning strategy by training across multiple tasks, enabling fast generalization to novel scenarios with minimal data. For instance, model-agnostic meta-learning (MAML)~\cite{Finn_2017_ICML} and optimization-based meta-learning~\cite{Ravi_2017_ICLR} have demonstrated fast convergence and strong generalization in few-shot and multi-task learning. These meta-learning methods provide robots with enhanced adaptability and decision-making efficiency in complex and dynamic environments. O'Connell et al.~\cite{O_2022_SR} demonstrated rapid learning and highly adaptive flight under different wind conditions. Their approach leveraged NN and online learning strategies to model the NN via meta-learning, effectively capturing external disturbance information that affects drone performance. In addition, an adaptive controller was designed to help the drone adjust to varying wind conditions, achieving robust flight control. This work offers an effective solution for autonomous flight in complex wind conditions. Building on this, Xie et al.~\cite{Xie_2024_ICRA} proposed a hierarchical meta-learning framework for drone control that extends the work in~\cite{O_2022_SR}. Their method decomposes tasks into multi-level sub-tasks and applies meta-learning at each level. Both studies highlight the potential of meta-learning for achieving adaptive and robust control. Although the learning algorithm in~\cite{O_2022_SR} combines meta-learning with an adaptive controller, it relies on only one trajectory in the training dataset because the NN is highly dependent on the trajectory, which may lead to poor generalization in robotic manipulator tasks.

In physical HRI, the human does not play a fixed role. Instead, the human may either actively guide the robot or passively follow its motion. These two interaction modes lead to different control objectives. In the active mode, the controller should track the human-applied force, whereas in the passive mode, the controller should compensate for the human disturbance to ensure stable and accurate motion execution. However, most existing approaches rely on predefined switching thresholds or external force sensors to distinguish modes and obtain force information. Moreover, pure data-driven approaches lack stability guarantees, while model-based controllers (MBCs) have limited adaptability. These limitations motivate a unified framework that integrates mode recognition with adaptive learning-based control to achieve both adaptability and stability in HRI.

The overview of the proposed work is illustrated in Fig.\ref{fig:Diagram}. The proposed framework integrates meta-learning based mode recognition and adaptive meta-learning robot control, and it is experimentally validated. The framework employs a meta-learning-trained NN to classify human intention, allowing more stable adaptation to new trajectories. A multiple-trajectory algorithm is developed to train the environment-basis NN and the force-dependent parameters. In addition, an adaptive meta-learning controller is designed to ensure the robustness of manipulator tracking during the human passive mode.

In summary, compared with existing literature in \cite{Bai_2021_TRO}, \cite{Chen_2024_TIE}, \cite{Chen_2025_IECON}, and \cite{Pollayil_2023_TRO}, this paper presents the following contributions:

\begin{enumerate}
  \item Compared with \cite{Bai_2021_TRO} where switching thresholds are manually defined, a meta-learning-based mode recognition method is proposed to classify human intention using an NN without thresholds. Compared with conventional NN training, meta-learning enables more stable adaptation when encountering new trajectories.
  
  \item Compared with the single trajectory case in \cite{Chen_2025_IECON}, a new algorithm is developed to train the environment-basis NN from multiple trajectories, allowing the controller to effectively capture unmodeled dynamics.

  \item An adaptive meta-learning controller is designed to compensate for unknown external disturbances and achieves more accurate tracking performance on unseen trajectories compared with controllers in \cite{Chen_2024_TIE} and \cite{Pollayil_2023_TRO}. The controller integrates data-driven learning with MBC, ensuring the convergence based on Lyapunov method.

  \item The proposed framework establishes a coupling between control and mode recognition. The recognition accuracy improves by incorporating the estimated force-dependent parameters into the mode recognition network. Experimental results in simulation and real robot demonstrate the effectiveness of the control performance and mode recognition in the proposed framework.
\end{enumerate}

\section{Robot Manipulator Dynamics}
As shown in Fig.\ref{fig:Diagram}, the manipulator used in this study is a 7-DOF Franka Emika robot, which provides an interface for joint torque control. The robot's joints operate with a signal transmission frequency of 1 kHz. The dynamics of the manipulator in joint space are expressed as:
\begin{equation}
    \mathcal{M}(\bm{q})\bm{\ddot{q}}+\mathcal{C}(\bm{q},\bm{\dot{q}})\bm{\dot{q}}+\mathcal{G}(\bm{q})=\bm{\tau}+\bm{\tau}_{ext},\label{dynamic_js}
\end{equation}
where $\mathcal{M}(\bm{q}) \in \mathbb{R}^{7\times7}$ denotes the inertia matrix, $\mathcal{C}(\bm{q}, \bm{\dot{q}}) \in \mathbb{R}^{7\times7}$ represents the Coriolis and centripetal matrix, and $\mathcal{G}(\bm{q}) \in \mathbb{R}^{7}$ is the gravity vector. $\bm{\tau} \in \mathbb{R}^{7}$ denotes the control torque input vector, and $\bm{\tau}_{ext} \in \mathbb{R}^{7}$ represents the combined external disturbances arising from internal friction, unmodeled dynamics, and external forces. \bm{$q$}, \bm{$\dot{q}$}, and $\bm{\ddot{q}} \in \mathbb{R}^{7}$ are the joint position, velocity, and acceleration vectors, respectively.

Inspired by~\cite{O_2022_SR}, we assume that the external disturbance term $\bm{\tau}_{ext}$ can be approximated as
\begin{equation}
    \bm{\tau}_{ext} \approx \phi(\bm{q}, \bm{\dot{q}})\bm{a},
\end{equation}
where $\phi$ is an NN that captures the unmodeled dynamics of the robot states and is shared across all environmental conditions. The vector $\bm{a} \in \mathbb{R}^{7}$ is the linear parameters specific to each environment. In the context of the HRI task, $\bm{a}$ corresponds to the parameter associated with the human force $F$.

This formulation allows the disturbance to be learned offline while enabling efficient online adaptation through parameter $\bm{a}$. The proposed representation integrates the advantages of data-driven learning and adaptive control. NN captures environment disturbance features from data, while the parameter $\bm{a}$ enables real-time adjustment of the controller. Compared with classical linear parameterization $\bm{\tau}_{ext} = Y(\bm{q}, \bm{\dot{q}}, \bm{\dot{q}}_d, \bm{\ddot{q}}_d)\bm{\theta}$ \cite{Spong_2020_Book}, this formulation improves approximation capability. Compared with pure NN model $\bm{\tau}_{ext} \approx f_{\bm{\theta}}(\bm{q}, \bm{\dot{q}})$ \cite{Li_2024_Processes}, it preserves a linear parameter structure, which is essential for adaptive control and Lyapunov-based convergence analysis. Gaps between the simulation and real environments are partially captured by the parameter $\bm{a}$ and compensated online through the adaptive law, thereby reducing the sim-to-real gap.
\begin{assumption} \label{as1}
    The optimal disturbance parameter $\bm{a}^*$ is first-order differentiable and there exist positive constants $\bar a$ and $d_a$, such that $\|\bm{a}^*\|\le \bar a,$ and $ \|\bm{\dot a}^*\|\le d_a$ are globally bounded.
\end{assumption}

\section{Meta-Learning Algorithm Design}
\subsection{Meta-Learning for Mode Recognition}
Conventional mode recognition control methods typically rely on manually defined thresholds for mode switching. However, such approaches are highly dependent on the specific task type and the individual characteristics of participants, which limits their generalization capability. In contrast, NNs can learn from multi-subject and multi-task datasets to extract broadly applicable representations, thereby eliminating the need for manually defined thresholds.

Although conventional training methods allow NNs to achieve strong performance on training datasets, they often fail to maintain comparable performance when evaluated on unseen test data. In particular, in HRI scenarios, data collection is both complex and costly, with force-related interaction data being especially scarce. These challenges place greater demands on the generalization and adaptability of NNs.

Therefore, we introduce model-agnostic meta-learning for mode recognition, a meta-learning algorithm proposed by Finn et al.~\cite{Finn_2017_ICML}. The objective of MAML is to learn a set of initial parameters $\theta$ that enable rapid adaptation to any new task sampled from a related distribution. Specifically, during each training iteration, a batch of tasks $\mathcal{D}_i$ is sampled from the task distribution $p(\mathcal{D})$. For each task $\mathcal{D}_i$, the dataset is divided into two subsets: an adaptation set and a training set. A few gradient descent steps are then performed on the adaptation set to obtain task-specific parameters:
\begin{equation}
    \theta^{'}_i = \theta - \eta_{in} \nabla_\theta \mathcal{L}_{\mathcal{D}_i}(f_\theta),\label{theta_dot}
\end{equation}
where $\eta_{in}$ denotes the inner-loop learning rate, and $\mathcal{L}_{\mathcal{D}i}(f_\theta)$ represents the loss computed on the adaptation set of task $\mathcal{D}_i$. \eqref{theta_dot} simulates how the model adapts to a new task starting from the shared initialization $\theta$.

After obtaining the adapted parameters $\theta_i^{'}$ for each task $\mathcal{D}_i$, these parameters are evaluated on the corresponding training set to compute the meta-loss $\mathcal{L}_{\mathcal{D}_i}(f_{\theta_i^{'}})$. The original parameters $\theta$ are then updated by performing backpropagation through the meta-losses as follows:
\begin{equation}
    \theta \gets \theta - \eta_{out} \nabla_\theta \sum_{\mathcal{D}_i \sim p(\mathcal{D})} \mathcal{L}_{\mathcal{D}_i}(f_{\theta^{'}_i}),
\end{equation}
where $\eta_{out}$ denotes the outer-loop learning rate.

\RestyleAlgo{ruled}
\SetKwComment{Comment}{/* }{ */}
\begin{algorithm}[t]
\caption{Proposed Multiple-Trajectory Meta-Learning (MTML) Algorithm}\label{alg:Training}
\textbf{Input}: $\mathcal{D}^{\phi}$\\
\textbf{Output}: Trained neural network $\phi$\\
\textbf{Initialize}: Neural network $\phi$ and auxiliary network $t$\\
\textbf{Hyperparameter}: $\rho_1, \rho_2, \bar{a}, \delta > 0$\\
\While{not done}{
    Randomly sample a batch of tasks $\mathcal{D}_{ij}^{\phi}\sim p(\mathcal{D}^{\phi})$\;
    \For{each task $\mathcal{D}_{ij}^{\phi}$}{
        Randomly sample disjoint adaptation and training set:\\
        $\prescript{a}{}{\mathcal{D}_{ij}^{\phi}}=\{\prescript{a}{}{\bm{x}}_{ij}^{\phi},\prescript{a}{}{\bm{y}}_{ij}^{\phi}\}$, $\prescript{t}{}{\mathcal{D}_{ij}^{\phi}=\{\prescript{t}{}{\bm{x}}_{ij}^{\phi}},\prescript{t}{}{\bm{y}}_{ij}^{\phi}\}$\;
        Solve $\hat{\bm{a}}_{ij}$ using $\prescript{a}{}{\mathcal{D}_{ij}^{\phi}}$:\\
        $\hat{\bm{a}}_{ij}=\arg\min_{\bm{a}} \sum \left\|\prescript{a}{}{\bm{y}}_{ij}^{\phi} - \phi(\prescript{a}{}{\bm{x}}_{ij}^{\phi}) \bm{a}\right\|^2$\;
        Compute mean parameters $\bar{\bm{a}}_{j}$ for the same force condition with $\hat{\bm{a}}_{ij}$\;
        Update $\bm{a}_{ij}$ with $\hat{\bm{a}}_{ij}$ and $\bar{\bm{a}}_{j}$:$~~~~~~~~~~~~~~~~${
        $\bm{a}_{ij} \gets \bm{a}_{ij} + \rho_1 (\hat{\bm{a}}_{ij} - \bm{a}_{ij}) + \rho_2 (\bar{\bm{a}}_{j} - \bm{a}_{ij})$
        }\;
        Normalize $\bm{a}_{ij}$: $\bm{a}_{ij} \gets \bm{a}_{ij} + \bar{a} \frac{\bm{a}_{ij}}{||\bm{a}_{ij}||}, \text{if}~||\bm{a}_{ij}|| > \bar{a}$\;
    }
    Update network $\phi$ via Adam optimizer using the loss:
    $\sum_{\prescript{t}{}{\mathcal{D}_{ij}^{\phi}}} 
    \left( 
    \|\prescript{t}{}{\bm{y}}_{ij}^{\phi} - \phi(\prescript{t}{}{\bm{x}}_{ij}^{\phi}) \bm{a}_{ij}\|^2
    - \delta \cdot \mathrm{loss}(t(\phi(\prescript{t}{}{\bm{x}}_{ij}^{\phi})), i)
    \right)$\;
    Update network $t$ via Adam optimizer using the loss:
    $\sum_{\prescript{t}{}{\mathcal{D}_{ij}^{\phi}}} 
    \mathrm{loss}(t(\phi(\prescript{t}{}{\bm{x}}_{ij}^{\phi})), i)$\;
}
\end{algorithm}
\subsection{Proposed Multiple-Trajectory Meta-Learning Algorithm}
In~\cite{O_2022_SR}, domain adversarially invariant meta-learning (DAIML) was proposed to model all wind-invariant disturbances affecting a drone using an NN, while representing wind-specific effects with a linear coefficient $\bm{a}^*$, which is updated in real time through adaptive control. Compared with conventional meta-learning, which updates all network parameters, this approach enables faster system response by adapting only the coefficient $\bm{a}^*$, thereby reducing computational requirements and improving real-time performance. Moreover, by incorporating an adaptive controller, the method simultaneously reduces control error and enhances system robustness and overall performance. In~\cite{Chen_2025_IECON}, a training algorithm inspired by DAIML was proposed to address the abrupt change problem of $\bm{a}^*$ and improve the convergence speed of the network. However, although both methods can handle different environments, they each consider only one trajectory in the training dataset. Training with a single trajectory limits generalization so multiple trajectories are required for effective manipulator network training.

One challenge in training a multiple-trajectory dataset is that the parameter $\bm{a}^*$ may be highly dependent on the trajectory type. This dependence is reasonable, as the network $\phi(\bm{q}, \dot{\bm{q}})$ is naturally influenced by the trajectory. To ensure that the parameter $\bm{a}^*$ across different trajectories and force conditions depends only on the applied force, we propose a multiple-trajectory meta-learning (MTML) algorithm to train the network $\phi(\bm{q}, \dot{\bm{q}})$. In the proposed algorithm, the update rule for $\bm{a}_{ij}$ balances data fitting and intra-class consistency. Specifically, the term $\hat{\bm{a}}_{ij}$ minimizes the fitting error for each trajectory, while the term $\bar{\bm{a}}_j$ enforces consistency among samples under the same force condition. This design reduces trajectory-dependent features and this decoupling is encouraged through regularization rather than strictly guaranteed. 

An auxiliary network $t$ is trained to further remove the correlation between the trajectory types and the NN $\phi$. The input of the auxiliary network $t$ is the feature representation generated by the NN $\phi$, and the output corresponds to the predicted trajectory index. The auxiliary network $t$ predicts the trajectory index $i$ from $\phi(\prescript{t}{}{\bm{x}}_{ij}^{\phi})$ using a cross-entropy loss. Through adversarial training, $t$ is optimized to distinguish different trajectories, while $\phi$ is optimized to prevent $t$ from identifying the trajectory. Therefore, trajectory-specific information is progressively suppressed in $\phi$, encouraging the learned representation to capture features shared across trajectories and improving its generalization to unseen trajectories.

\section{Adaptive Meta-Learning Controller Design}
\subsection{Proposed Controller Design}
The feedback control component of the proposed controller is inspired by the non-singular terminal sliding mode (NTSM) controller presented in~\cite{Feng_2002_Automatica}. The sliding surface is defined as
\begin{equation}
    \bm{s} = \bm{e} + \beta \bm{\dot{e}}^{\alpha}, \label{s}
\end{equation}
where $\beta > 0$, $\alpha = p/q$, and $p > 0$ and $q > 0$ are adjacent odd numbers such that $1 < \alpha < 2$. The position and velocity errors are defined as $\bm{e} = \bm{q} - \bm{q}_d \in \mathbb{R}^7$ and $\dot{\bm{e}} = \dot{\bm{q}} - \dot{\bm{q}}_d \in \mathbb{R}^7$, respectively. The derivative of the sliding surface \eqref{s} can then be expressed as
\begin{equation}
    \dot{\bm{s}} = \bm{\dot{e}} + \alpha \beta \bm{\dot{e}}^{\alpha-1} \bm{\ddot{e}}.\label{s_dot_1}
\end{equation}
    
The proposed adaptive meta-learning (AML) controller can be summarized by the following control and update laws:
\begin{align}
    \bm{\tau}_{\textrm{AML}} = & \underbrace{\mathcal{M}(\bm{q}) \bm{\ddot{q}}_{d} + \mathcal{C}(\bm{q},\bm{\dot{q}})\bm{\dot{q}} + \mathcal{G}(\bm{q})}_{\text{feedforward}} \underbrace{ - \Phi \bm{\hat{a}}}_{\text{meta-learning term}} \nonumber\\
    & \underbrace{+ \mathcal{M}(\bm{q}) \Big( \frac{-\bm{\dot{e}}^{2-\alpha}}{\alpha \beta} - K_1 \textrm{tanh}(\kappa \bm{s}) \Big) - K_2 \textrm{tanh}(\kappa \bm{s})}_{\text{feedback}},\\
    \dot{\hat{\bm{a}}} = & \alpha \beta \bm{\dot{e}}^{\alpha-1} \mathcal{M}(\bm{q})^{-1} \Gamma \Phi^T \bm{s} - \sigma \bm{\hat a}, \label{update_law}
\end{align}
where $\bm{\tau}_{\textrm{AML}}$ denotes the control torque input, $\dot{\hat{\bm{a}}}$ represents the parameter update law, $\sigma$ is a positive regularization gain, and $\Gamma$ is a diagonal positive-definite adaptation gain matrix. $\Gamma$ determines the update rate of $\hat{\bm{a}}$. In the update law \eqref{update_law}, a larger $\Gamma$ leads to a faster adaptation but may increase the sensitivity, while a smaller $\Gamma$ results in a smoother but slower update. The vector $\bm{\ddot{q}}_d \in \mathbb{R}^{7}$ represents the desired joint acceleration. The output of the NN $\phi(\bm{q}, \bm{\dot{q}})$ is a 7-dimensional vector. For ease of computation in the control law, the output of $\phi$ is reorganized into a diagonal matrix $\Phi = \operatorname{diag}\{\phi(\bm{q}, \bm{\dot{q}})\} \in \mathbb{R}^{7 \times 7}$ to preserve the linear parameter structure and facilitate convergence analysis.

The continuous $\textrm{tanh}(\kappa \bm{s})$ function is employed in place of the discontinuous $\textrm{sgn}(\bm{s})$ function, where $\kappa > 0$ is a design parameter. This smoothing function mitigates chattering near the sliding surface. In our previous work~\cite{Chen_2024_TIE} and~\cite{Wan_2025_TIE}, the controller was designed in the task space. However, in the joint space, the mass matrix $\mathcal{M}(\bm{q})$ significantly reduces the control effect on the last three joints of the manipulator. Therefore, an additional feedback term $-K_2 \textrm{tanh}(\kappa \bm{s})$ is introduced to compensate for this effect.

\subsection{Convergence Analysis}
The closed-loop dynamics can be obtained by substituting the proposed controller into the system dynamics in \eqref{dynamic_js} with the following expression:
\begin{align}
    \ddot{\bm{e}} &= \frac{-\bm{\dot{e}}^{2-\alpha}}{\alpha \beta} - K_1 \textrm{tanh}(\kappa \bm{s}) \nonumber\\
    & \quad - \mathcal{M}(\bm{q})^{-1} \Big(K_2 \textrm{tanh}(\kappa \bm{s}) + \Phi \tilde{\bm{a}}\Big),\label{e_ddot}
\end{align}
where parameter estimation error $\tilde{\bm{a}}$ is defined as $\tilde{\bm{a}} = \hat{\bm{a}} - \bm{a}^*$.Substituting \eqref{e_ddot} into the derivative of the sliding surface \eqref{s_dot_1},
\begin{align}
    \dot{\bm{s}} &= \bm{\dot{e}} + \alpha \beta \bm{\dot{e}}^{\alpha-1} \bm{\ddot{e}}, \nonumber\\
    &= \bm{\dot{e}} + \alpha \beta \bm{\dot{e}}^{\alpha-1} \Big[\frac{-\bm{\dot{e}}^{2-\alpha}}{\alpha \beta} - K_1 \textrm{tanh}(\kappa \bm{s}) \nonumber\\
    & \quad - \mathcal{M}(\bm{q})^{-1} \Big(K_2 \textrm{tanh}(\kappa \bm{s}) + \Phi \tilde{\bm{a}}\Big) \Big], \nonumber\\
    &= \bm{\dot{e}} - \bm{\dot{e}}^{\alpha-1}\bm{\dot{e}}^{2-\alpha} - \alpha \beta \bm{\dot{e}}^{\alpha-1} K_1 \textrm{tanh}(\kappa \bm{s}) \nonumber \\
    & \quad - \alpha \beta \bm{\dot{e}}^{\alpha-1} \mathcal{M}(\bm{q})^{-1} \Big(K_2 \textrm{tanh}(\kappa \bm{s}) +\Phi \tilde{\bm{a}}\Big), \nonumber \\
    &= - \alpha \beta \bm{\dot{e}}^{\alpha-1} K_1 \textrm{tanh}(\kappa \bm{s}) \nonumber \\
    & \quad - \alpha \beta \bm{\dot{e}}^{\alpha-1} \mathcal{M}(\bm{q})^{-1} \Big(K_2 \textrm{tanh}(\kappa \bm{s}) +\Phi \tilde{\bm{a}}\Big).
    \label{s_dot}
\end{align}

Construct a Lyapunov function in a quadratic form as
\begin{equation}
    V = \frac{1}{2} \bm{s}^T \bm{s} + \frac{1}{2} \tilde{\bm{a}}^T \Gamma^{-1} \tilde{\bm{a}},\label{V}
\end{equation}
which jointly represents the tracking error and parameter estimation error. The term $V_s=\frac{1}{2}\bm{s}^T\bm{s}$ characterizes the sliding surface, while $V_a=\frac{1}{2}\tilde{\bm{a}}^T \Gamma^{-1}\tilde{\bm{a}}$ accounts for the parameter estimation error. The derivative of \eqref{V} is given by
\begin{align}
    \dot{V} = \bm{s}^T \bm{\dot{s}} + \tilde{\bm{a}}^T \Gamma^{-1} \dot{\tilde{\bm{a}}}. \label{v_dot_1}
\end{align}

Substituting \eqref{s_dot} into \eqref{v_dot_1}, we obtain
\begin{align}
    \dot{V} &= - \alpha \beta \bm{\dot{e}}^{\alpha-1} K_1 \bm{s}^T \textrm{tanh}(\kappa \bm{s}) \nonumber\\
    & \quad - \alpha \beta \bm{\dot{e}}^{\alpha-1} \mathcal{M}(\bm{q})^{-1} \Big(K_2 \bm{s}^T\textrm{tanh}(\kappa \bm{s}) +\bm{s}^T\Phi \tilde{\bm{a}}\Big)\nonumber\\
    & \quad + \tilde{\bm{a}}^T \Gamma^{-1} \dot{\tilde{\bm{a}}}\nonumber\\
    &= - \alpha \beta \bm{\dot{e}}^{\alpha-1} K_1 \bm{s}^T \textrm{tanh}(\kappa \bm{s}) \nonumber\\
    & \quad - \alpha \beta \bm{\dot{e}}^{\alpha-1} \mathcal{M}(\bm{q})^{-1} \Big(K_2 \bm{s}^T\textrm{tanh}(\kappa \bm{s}) +\bm{s}^T\Phi \tilde{\bm{a}}\Big)\nonumber\\
    & \quad + \tilde{\bm{a}}^T \Gamma^{-1} \Big(\alpha \beta \bm{\dot{e}}^{\alpha-1} \mathcal{M}(\bm{q})^{-1} \Gamma \Phi^T \bm{s} - \sigma \bm{\tilde a} - \sigma \bm{a}^* -\dot{\bm{a}}^*\Big)\nonumber\\
\end{align}
\begin{align}
    &= - \alpha \beta \bm{\dot{e}}^{\alpha-1} \Big(K_1 + K_2 \mathcal{M}(\bm{q})^{-1}\Big) \bm{s}^T \textrm{tanh}(\kappa \bm{s}) \nonumber\\
    & \quad - \alpha \beta \bm{\dot{e}}^{\alpha-1} \mathcal{M}(\bm{q})^{-1} \bm{s}^T \Phi \tilde{\bm{a}}\nonumber\\
    & \quad +  \alpha \beta \bm{\dot{e}}^{\alpha-1} \mathcal{M}(\bm{q})^{-1} \tilde{\bm{a}}^T \Phi^T \bm{s}\nonumber\\
    & \quad -\sigma \tilde{\bm{a}}^T\Gamma^{-1}\tilde{\bm{a}} -\sigma \tilde{\bm{a}}^T\Gamma^{-1}\bm{a}^* - \tilde{\bm{a}}^T\Gamma^{-1}\bm{\dot a}^*.\label{v_dot_2}
\end{align}

Since $\Phi$ is a diagonal matrix and both $\tilde{\bm{a}}$ and $\bm{s}$ are vectors in $\mathbb{R}^7$, the scalar products $\bm{s}^T \Phi \tilde{\bm{a}}$ and $\tilde{\bm{a}}^T \Phi^T \bm{s}$ are equivalent. Then \eqref{v_dot_2} can be rewritten as
\begin{equation}
    \dot{V} = -W(\bm{s}) -\sigma \tilde{\bm{a}}^T\Gamma^{-1}\tilde{\bm{a}} -\sigma \tilde{\bm{a}}^T\Gamma^{-1}\bm{a}^* - \tilde{\bm{a}}^T\Gamma^{-1}\bm{\dot a}^*,
\end{equation}
where
\begin{equation}
    W(s)= \alpha\beta \bm{\dot e}^{\alpha-1}\bm{s}^T\left(K_1+K_2M(\bm{q})^{-1}\right)\tanh(\kappa \bm{s}).
\end{equation}

Using the Cauchy--Schwarz inequality together with Assumption \ref{as1} gives
\begin{equation}
    \dot V \le -W(\bm{s}) -\sigma \tilde{\bm{a}}^T\Gamma^{-1}\tilde{\bm{a}} +\|\Gamma^{-1}\|(\sigma \bar a+d_a)\|\tilde{\bm{a}}\|.
\end{equation}

Using $\tilde{\bm{a}}^T\Gamma^{-1}\tilde{\bm{a}} \ge \lambda_{\min}(\Gamma^{-1})\|\tilde{\bm{a}}\|^2$, it follows that 
\begin{equation}
    \dot V \le -W(\bm{s}) -\sigma\lambda_\Gamma\|\bm{\tilde a}\|^2 +c_0\|\bm{\tilde a}\|,\label{v_dot_3}
\end{equation}
where $\lambda_\Gamma=\lambda_{\min}(\Gamma^{-1}), c_0=\|\Gamma^{-1}\|(\sigma \bar a+ d_a)$. Applying Young's inequality $ c_0\|\bm{\tilde a}\| \le \frac{\sigma\lambda_\Gamma}{2}\|\bm{\tilde a}\|^2 + \frac{c_0^2}{2\sigma\lambda_\Gamma}$, \eqref{v_dot_3} can be rearranged as
\begin{equation}
    \dot V \le -W(\bm{s}) -\frac{\sigma\lambda_\Gamma}{2}\|\bm{\tilde a}\|^2 +\Delta,
\end{equation}
where
\begin{equation}
    \Delta= \frac{c_0^2} {2\sigma\lambda_\Gamma} > 0.
\end{equation}

Define the scalar function
\begin{equation}
    f(x)=
    \begin{cases}
    \dfrac{x\tanh(\kappa x)}{x^2}, & x\neq 0,\\[6pt]
    \kappa, & x=0.
    \end{cases}
\end{equation}

Since \(f(x)\) is continuous and strictly positive on the compact set
\([-\bar s,\bar s]\), there exists a constant $c_{\bar s}=\min_{|x|\le \bar s} f(x) >0$ such that
$x\tanh(\kappa x)\ge c_{\bar s}x^2, \forall |x|\le \bar s$. Consequently, for all \(\bm{s}\le \bm{\bar s}\), $\bm{s}^T\tanh(\kappa \bm{s})\ge c_{\bar s}\|\bm{s}\|^2$. Considering that $\alpha$, $\beta$, $K_1$, $K_2$, $\bm{\dot{e}}^{\alpha-1}$, and $\bm{s}^T \textrm{tanh}(\kappa \bm{s})$ are all nonnegative, and that $\mathcal{M}(\bm{q})^{-1}$ is a positive-definite matrix, we can find positive values $\lambda_K$ and $d$ such that 
$K_1+K_2M^{-1}(\bm{q})\ge \lambda_K I, \bm{\dot e}^{\alpha-1}\ge d>0$, it follows that $W(\bm{s})\ge c_s\|\bm{s}\|^2$, where $c_s=\alpha\beta d\,\lambda_K\,c_{\bar s}>0$ and therefore
\begin{equation}
    \dot V \le -c_s\|\bm{s}\|^2 -\frac{\sigma\lambda_\Gamma}{2}\|\bm{\tilde a}\|^2 +\Delta.
\end{equation}

Since
$
V = \frac{1}{2} \bm{s}^T \bm{s} + \frac{1}{2} \tilde{\bm{a}}^T \Gamma^{-1} \tilde{\bm{a}} \le \frac12\|\bm{s}\|^2+\frac12\|\Gamma^{-1}\|\|\bm{\tilde a}\|^2,
$
there exists $\lambda=\min\left\{2c_s,\frac{\sigma\lambda_\Gamma}{\|\Gamma^{-1}\|}\right\}>0
$
such that $c_s\|\bm{s}\|^2+\frac{\sigma\lambda_\Gamma}{2}\|\bm{\tilde a}\|^2\ge \lambda V$.
Therefore,
\begin{equation}
    \dot V\le -\lambda V+\Delta.\label{v_dot_4}
\end{equation}

According to the Comparison Lemma \cite{Khalil_2002_Book}, \eqref{v_dot_4} leads to
\begin{equation}
    V(t)\le \left(V(0)-\frac{\Delta}{\lambda}\right)e^{-\lambda t}+\frac{\Delta}{\lambda}.
\end{equation}
Hence, the closed-loop system is uniformly ultimately bounded. The ultimate bounds can be obtained as

\begin{align}
    \limsup_{t\to\infty}V(t) &\le \frac{\Delta}{\lambda}\nonumber,\\
    \limsup_{t \to \infty} \|\bm{s}(t)\|
    &\le
    \sqrt{\frac{2\Delta}{\lambda}},\nonumber\\
    \limsup_{t \to \infty} \|\bm{\tilde{a}}(t)\|
    &\le
    \sqrt{\frac{2\Delta}{\lambda \lambda_{\Gamma}}}.\nonumber
\end{align}

This concludes the convergence analysis. \qed

\section{Performance Evaluation}
All experiments in this section are performed using a 7-DOF Franka Emika robotic manipulator without a gripper. The experiments are conducted on a standard desktop equipped with an Intel i7 CPU and Intel UHD Graphics 630 GPU. It is worth noting that all components in the proposed framework are executed in real time. Both the mode recognition network $\mathcal{N}$ and the environment-basis network $\bm{\phi}$ are integrated directly into the control loop. The total computation time is below 1\,ms, ensuring real-time performance without introducing additional control latency. To train and validate the performance of the NN and controller, 11 trajectories ($T_1\sim T_{11}$) are defined, where $T_1 \sim T_6$ are used for training and $T_7 \sim T_{11}$ for validation. All trajectories originate from the end-effector's initial position $[x_0, y_0, z_0]$. Trajectories $T_1 - T_8$ are defined as follows:
\begin{align}
     T_1 &=
    \left[\begin{matrix}
        0.1\operatorname{sin}(\frac{2\pi t}{5})\\
        - 0.1\operatorname{cos}(\frac{2\pi t}{5}) + 0.1\\
        0
    \end{matrix}\right],
    T_2 =
    \left[\begin{matrix}
        0.1\operatorname{sin}(\frac{2\pi t}{5})\\
        0\\
        - 0.1\operatorname{cos}(\frac{2\pi t}{5}) + 0.1
    \end{matrix}\right], \nonumber\\
    T_3 &=
    \left[\begin{matrix}
        0\\
        0.1\operatorname{sin}(\frac{2\pi t}{5})\\
        - 0.1\operatorname{cos}(\frac{2\pi t}{5}) + 0.1
    \end{matrix}\right],
    T_4 =
    \left[\begin{matrix}
        - 0.1\operatorname{sin}(\frac{2\pi t}{5})\\
        - 0.1\operatorname{cos}(\frac{2\pi t}{5}) + 0.1\\
        0
    \end{matrix}\right], \nonumber\\
    T_5 &=
    \left[\begin{matrix}
        - 0.1\operatorname{sin}(\frac{2\pi t}{5})\\
        0\\
        - 0.1\operatorname{cos}(\frac{2\pi t}{5}) + 0.1
    \end{matrix}\right],
    T_6 =
    \left[\begin{matrix}
        0\\
        - 0.1\operatorname{sin}(\frac{2\pi t}{5})\\
        - 0.1\operatorname{cos}(\frac{2\pi t}{5}) + 0.1
    \end{matrix}\right], \nonumber\\
    T_7 &=
    \left[\begin{matrix}
        0.1\operatorname{sin}(\frac{2\pi t}{5})\\
        - 0.1\operatorname{cos}(\frac{2\pi t}{5}) + 0.1\\
        - 0.1\operatorname{cos}(\frac{2\pi t}{5}) + 0.1
    \end{matrix}\right],
    T_8 =
    \left[\begin{matrix}
        0.15\operatorname{sin}(\frac{2\pi t}{1})\\
        - 0.15\operatorname{sin}(\frac{2\pi t}{1})\\
        - 0.1\operatorname{cos}(\frac{2\pi t}{1}) + 0.1
    \end{matrix}\right]. \nonumber\\ \nonumber
\end{align}
\begin{figure}[t]
    \centering
    \subfloat[]{
        \includegraphics[scale=0.045]{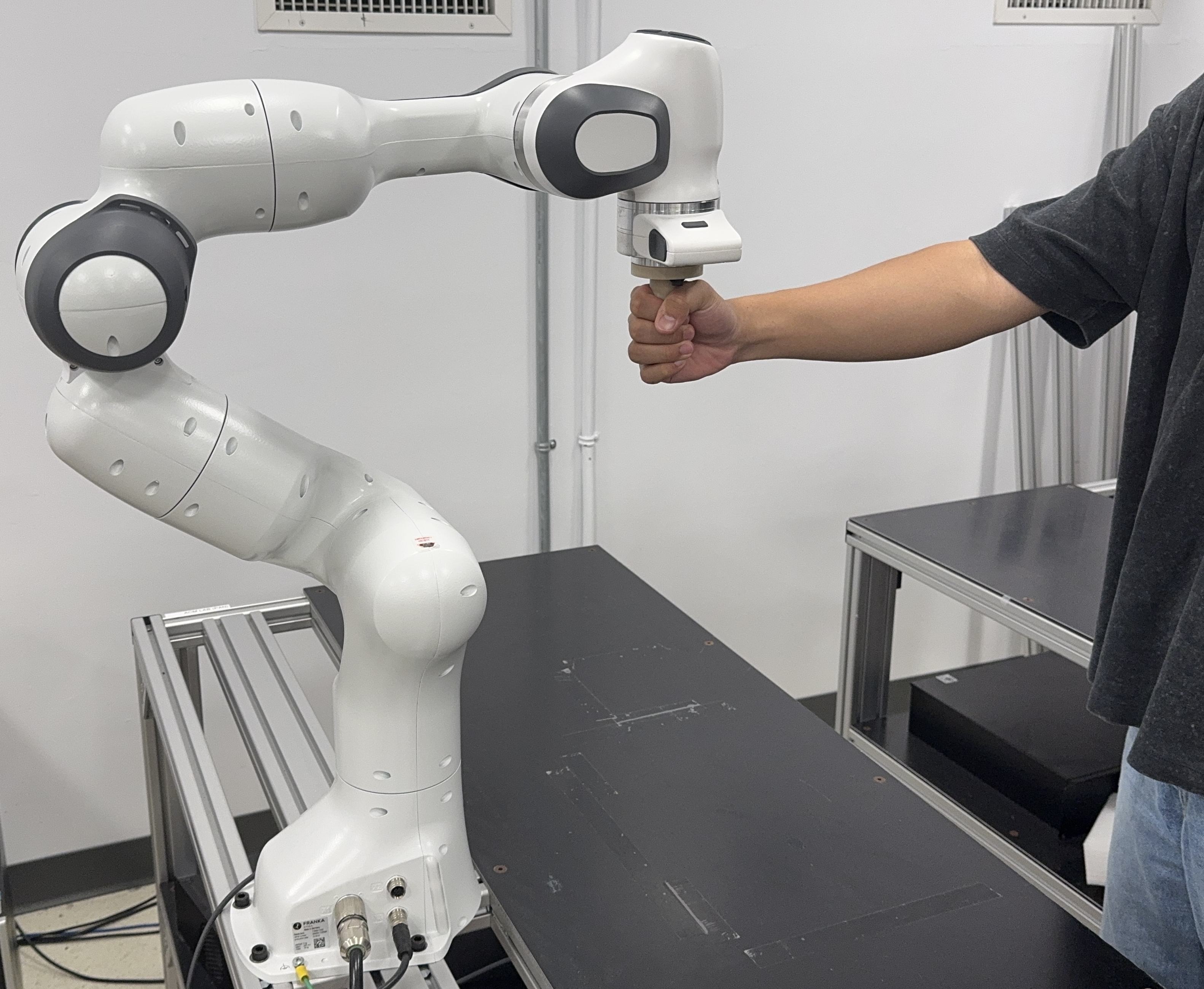}}
    \subfloat[]{
        \includegraphics[scale=0.185]{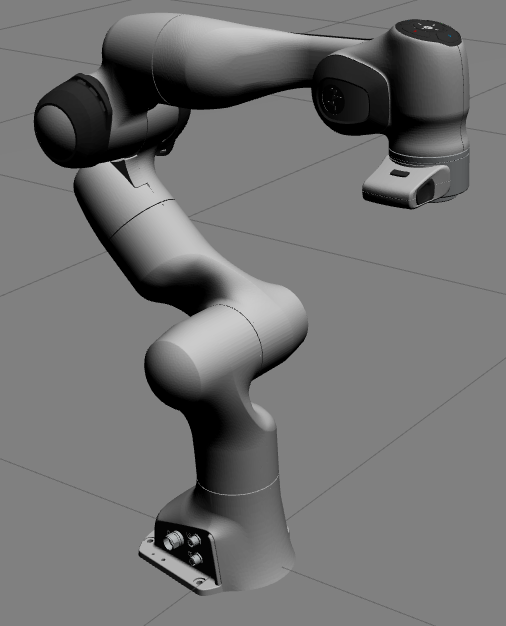}}
    \caption{Data collection environments for the Franka Emika robot. (a) Hardware setup with human interaction in the Advanced Control and Mechatronics Laboratory, Dalhousie University. (b) Gazebo simulation environment.}
    \label{fig:Simulation}
\end{figure}

Trajectories $T_9$ and $T_{11}$ are constant references updated every $2\,\mathrm{s}$ to evaluate the non-periodic tracking performance. Trajectory $T_{10}$ is an acceleration-deceleration motion, used to evaluate the tracking performance under variable-acceleration conditions.
\begin{figure*}[t]
    \centering
    \subfloat[]{
        \includegraphics[width=0.36\linewidth]{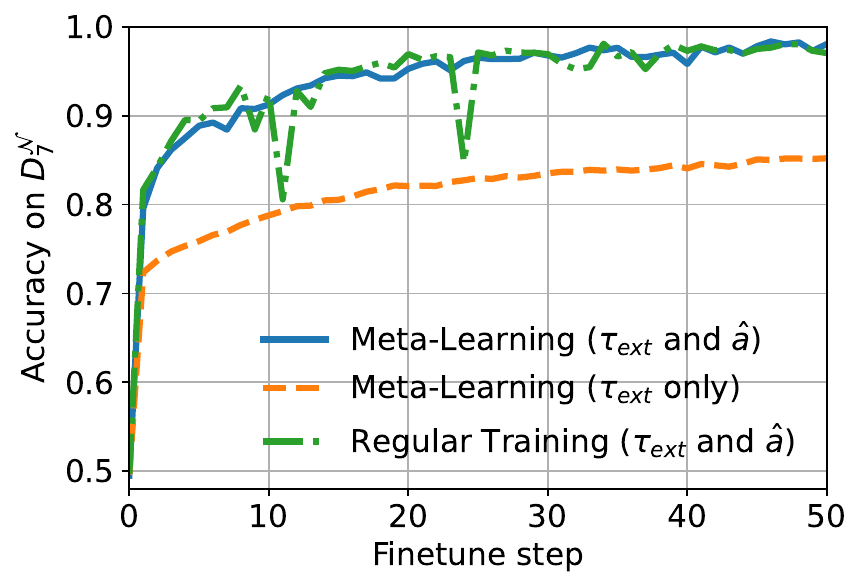}}
    \subfloat[]{
        \includegraphics[width=0.32\linewidth]{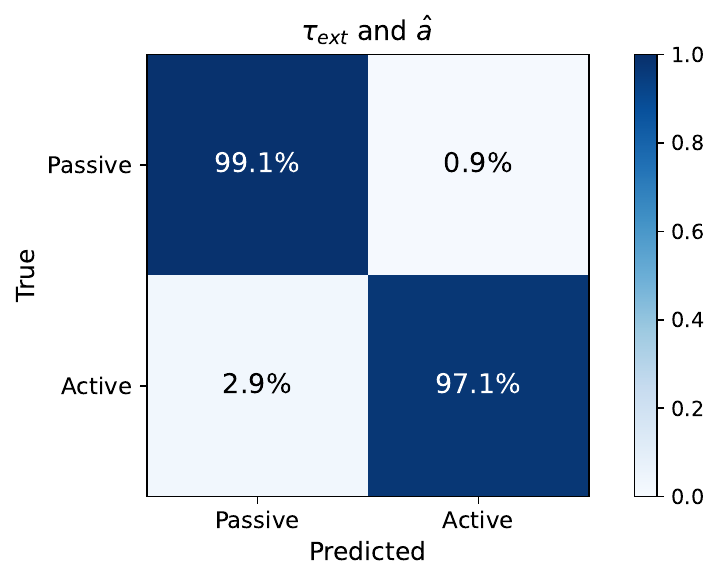}}
    \subfloat[]{
        \includegraphics[width=0.32\linewidth]{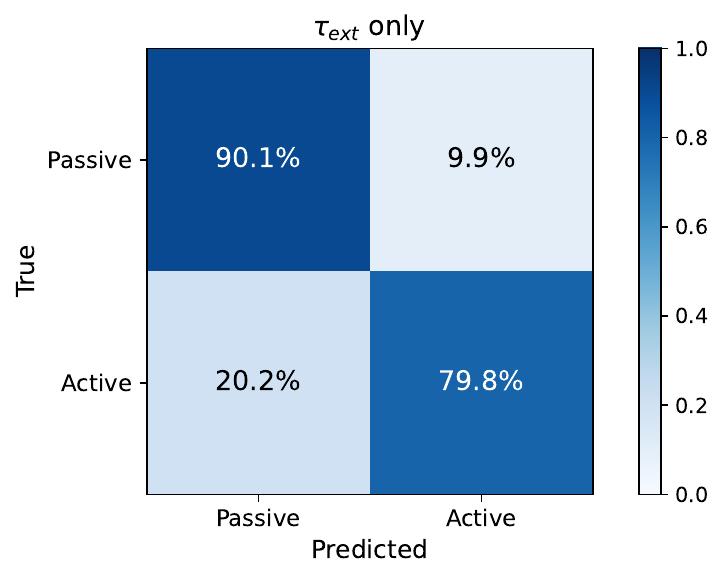}}
    \caption{Mode recognition NN $\mathcal{N}$ training results. (a) Adaptation comparison between the regular training, meta-learning with $\bm{\tau}_{ext}$ only, and meta-learning with both $\bm{\tau}_{ext}$ and $\bm{\hat{a}}$; (b) Confusion matrix of $\mathcal{N}(\bm{\tau}_{ext}, \bm{\hat{a}})$; (c) Confusion matrix of $\mathcal{N}(\bm{\tau}_{ext})$.}
    \label{fig:Mode_Recognition}
\end{figure*}
\begin{figure*}[t]
    \centering
    \subfloat[]{
        \includegraphics[width=0.33\linewidth]{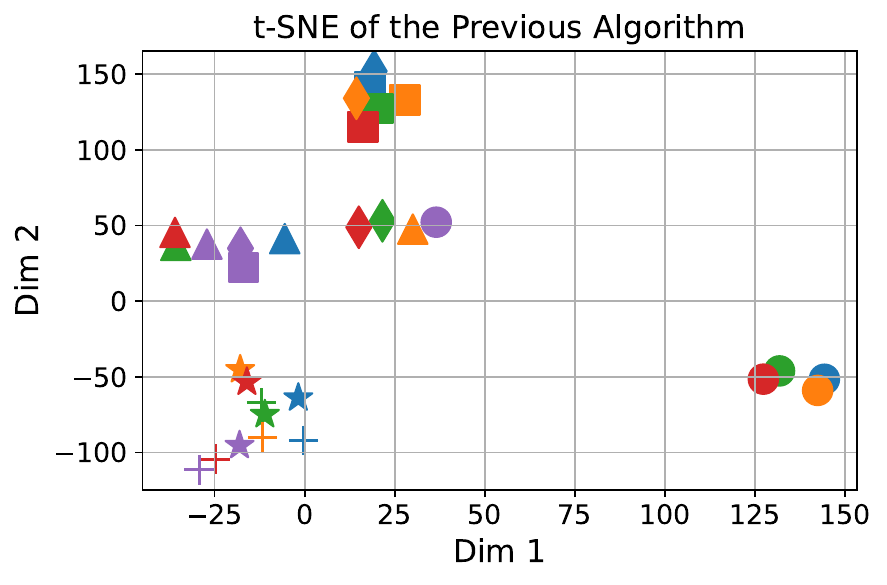}}
    \subfloat[]{
        \includegraphics[width=0.33\linewidth]{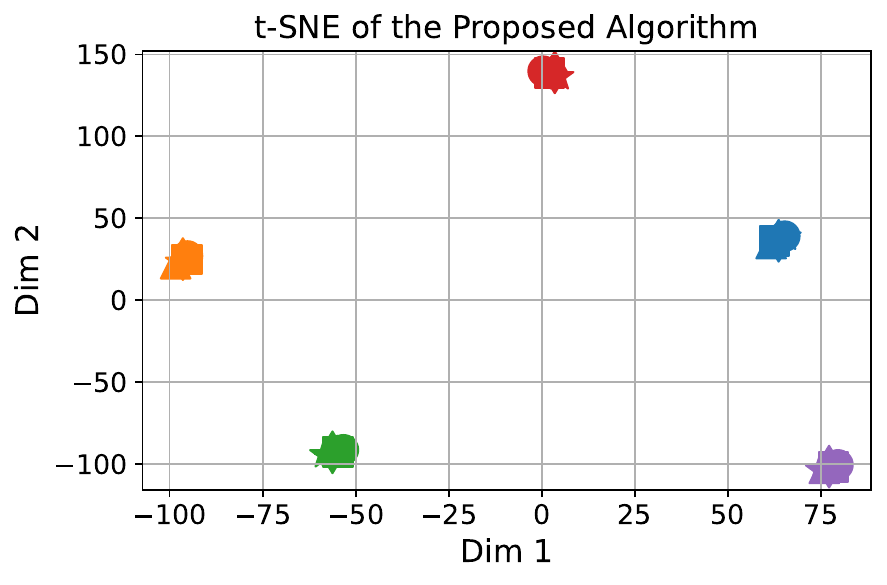}}
    \subfloat[]{
        \includegraphics[width=0.33\linewidth]{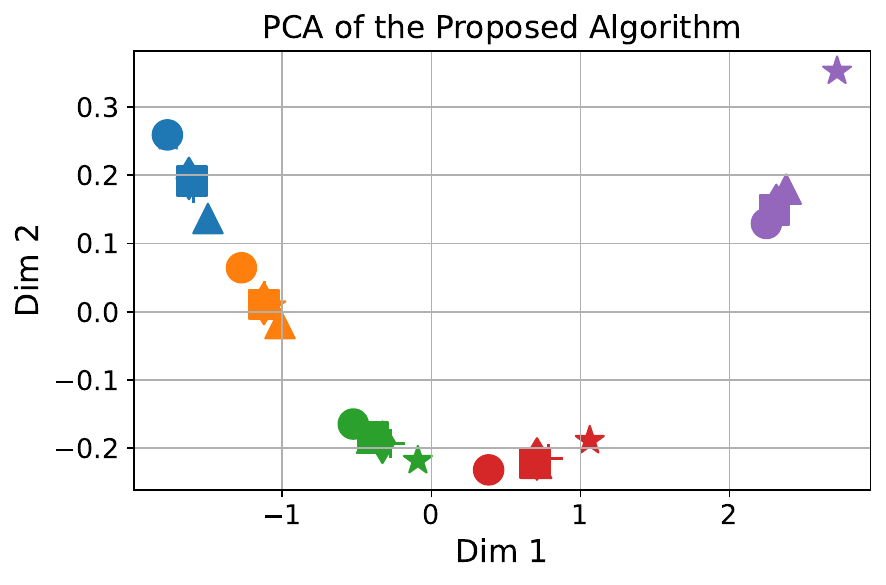}}
    \caption{Feature visualization of the parameter $\bm{a}$ under different trajectories and forces: (a) t-SNE visualization of $\bm{a}$ generated by the method in~\cite{Chen_2025_IECON}; (b) t-SNE visualization of $\bm{a}$ generated by the proposed MTML algorithm; (c) PCA visualization of $\bm{a}$ generated by the proposed MTML algorithm. Colors denote different force conditions $F_1 \sim F_5$, while marker shapes denote different trajectories $T_1 \sim T_6$ (circle, star, square, triangle, plus, and diamond, respectively).}
    \label{fig:Training_Comparison}
\end{figure*}

\subsection{Neural Network Training Details}
To train the mode recognition NN $\mathcal{N}$ using meta-learning, the manipulator is configured in the real world as shown in Fig.\ref{fig:Simulation} (a). The end-effector follows seven trajectories $T_1, T_2, \dots, T_7$, while five participants interact with it under both passive and active modes. All participants provided informed consent prior to participating in the experiments. Trajectories $T_1\sim T_6$ were used to train the NN $\mathcal{N}$, whereas $T_7$ was reserved for adaptation and evaluations. The participant involved in collecting the test dataset did not participate in the collection of the training data. $T_7$ active and passive recordings were independently and randomly divided into 20\% adaptation data and 80\% evaluation data. The NN was updated on the adaptation subset for 50 epochs using Stochastic gradient descent (SGD) with a learning rate of 0.01 and a batch size of 32.

The force-dependent parameter $\bm{\hat{a}}$ and the estimated external torque $\bm{\tau}_{ext}$ are collected as training inputs, while the passive and active modes serve as training outputs. $\bm{\tau}_{ext}$ is provided by the Franka Emika robot. It is obtained from the residual between the measured joint torque and the torque predicted by the robot's dynamic model, followed by low-pass filtering. Active recordings were assigned as label 1 and passive recordings as label 0 according to the experimental recording condition. The collected dataset is defined as $D^{\mathcal{N}} = {D^{\mathcal{N}}_1, D^{\mathcal{N}}_2, \dots, D^{\mathcal{N}}_{7}}$, where
\begin{equation}
    D^{\mathcal{N}}_i = \Bigl\{ \bm{x}^{\mathcal{N}}_i = [\bm{\hat{a}}, \bm{\tau}_{ext}],~\bm{y}^{\mathcal{N}}_i = \textrm{0 or 1}\Bigr\},\label{snn_dataset}\nonumber
\end{equation}
$\{D^{\mathcal{N}}_1, D^{\mathcal{N}}_2, \dots D^{\mathcal{N}}_{6}\}$ are training dataset, and $D^{\mathcal{N}}_{7}$ serves as the validation dataset. For each $D^{\mathcal{N}}_i$, the dataset contains approximately 25,000 samples.

To train the environment-basis NN $\phi$ for external disturbance compensation, the MTML algorithm requires the joint angle $\bm{q}$, joint velocity $\dot{\bm{q}}$, and the estimated external torque $\bm{\tau}_{ext}$. The manipulator is simulated in Gazebo, as shown in Fig.\ref{fig:Simulation} (b). The robot equipped with the proposed joint-space controller but without the learning feedback term, follows trajectories $T_1, T_2, \dots, T_6$, while five forces $F_1 = 0\textrm{N}$, $F_2 = 5\textrm{N}$, $F_3 = 10\textrm{N}$, $F_4 = 15\textrm{N}$, and $F_5 = 20\textrm{N}$ are applied to the end-effector with random directions, respectively. Each trajectory is sampled at 1~kHz. The collected dataset is defined as $D^{\phi} = {D^{\phi}_{11}, D^{\phi}_{12}, \dots, D^{\phi}_{15}, D^{\phi}_{21}, \dots, D^{\phi}_{65}}$, where
\begin{equation}
    D^{\phi}_{ij} = \Bigl\{ \bm{x}^{\phi}_{T_i, F_j} = [\bm{q}, \dot{\bm{q}}],~\bm{y}^{\phi}_{T_i, F_j} = [\bm{\tau}_{ext}] \Bigr\}. \label{mtml_dataset}\nonumber
\end{equation}

The NN $\mathcal{N}$ consists of two fully connected layers with a structure of $14 \to 32 \to 2$. The network $\phi$ consists of three fully connected layers with a structure of $14 \to 64 \to 32 \to 7$. The activation function used in each hidden layer is the hyperbolic tangent function. The entire learning framework is implemented in Python using PyTorch.

\subsection{Neural Network Training Results}
Fig.\ref{fig:Mode_Recognition} demonstrates the benefit of coupling the adaptive controller with the recognition network. When only the estimated external joint torque $\bm{\tau}_{ext}$ is used as the network input, the classifier correctly identifies 90.1\% of the passive samples and 79.8\% of the active samples. By additionally incorporating the adaptively estimated force-dependent parameter $\hat{\bm a}$, the correct classification rate increases to 99.1\% and 97.1\% for the passive and active modes, respectively. These results indicate that $\hat{\bm a}$ provides additional interaction information beyond $\bm{\tau}_{ext}$ and improves the separability between the two interaction modes. Moreover, the adaptation results on $D_7^{\mathcal{N}}$ show that the proposed meta-learning approach using both $\bm{\tau}_{ext}$ and $\hat{\bm a}$ achieves higher recognition accuracy during fine-tuning compared with using $\bm{\tau}_{ext}$ only. The regularly trained model with both inputs can eventually achieve a comparable accuracy, but exhibits unstable performance during fine-tuning, including noticeable temporary performance degradation. Overall, the results show that the adaptively estimated force-dependent parameter $\hat{\bm a}$ improves recognition capability, while meta-learning enables fast and more stable adaptation to previously unseen interaction data.

The t-distributed stochastic neighbor embedding (t-SNE) visualization of the parameter $\bm{a}$ compares the algorithm in~\cite{Chen_2025_IECON} with the proposed MTML algorithm. As shown in Fig.\ref{fig:Training_Comparison} (a), in the previous algorithm~\cite{Chen_2025_IECON}, the parameter $\bm{a}$ strongly correlates with the trajectory type, which is undesirable since $\bm{a}$ should be force-dependent only. In contrast, as illustrated in Fig.\ref{fig:Training_Comparison} (b), the proposed algorithm clusters parameters corresponding to the same force but different trajectories together, indicating that $\bm{a}$ is indeed force-dependent only. Principal components analysis (PCA) in Fig.\ref{fig:Training_Comparison} (c) demonstrates that samples corresponding to different force levels are separated along the principal components, indicating that the dominant variation in $\bm{a}$ is governed by the applied force rather than trajectory differences.

\subsection{Controller Results}
\begin{figure*}[t]
    \centering
    \includegraphics[width=1\linewidth]{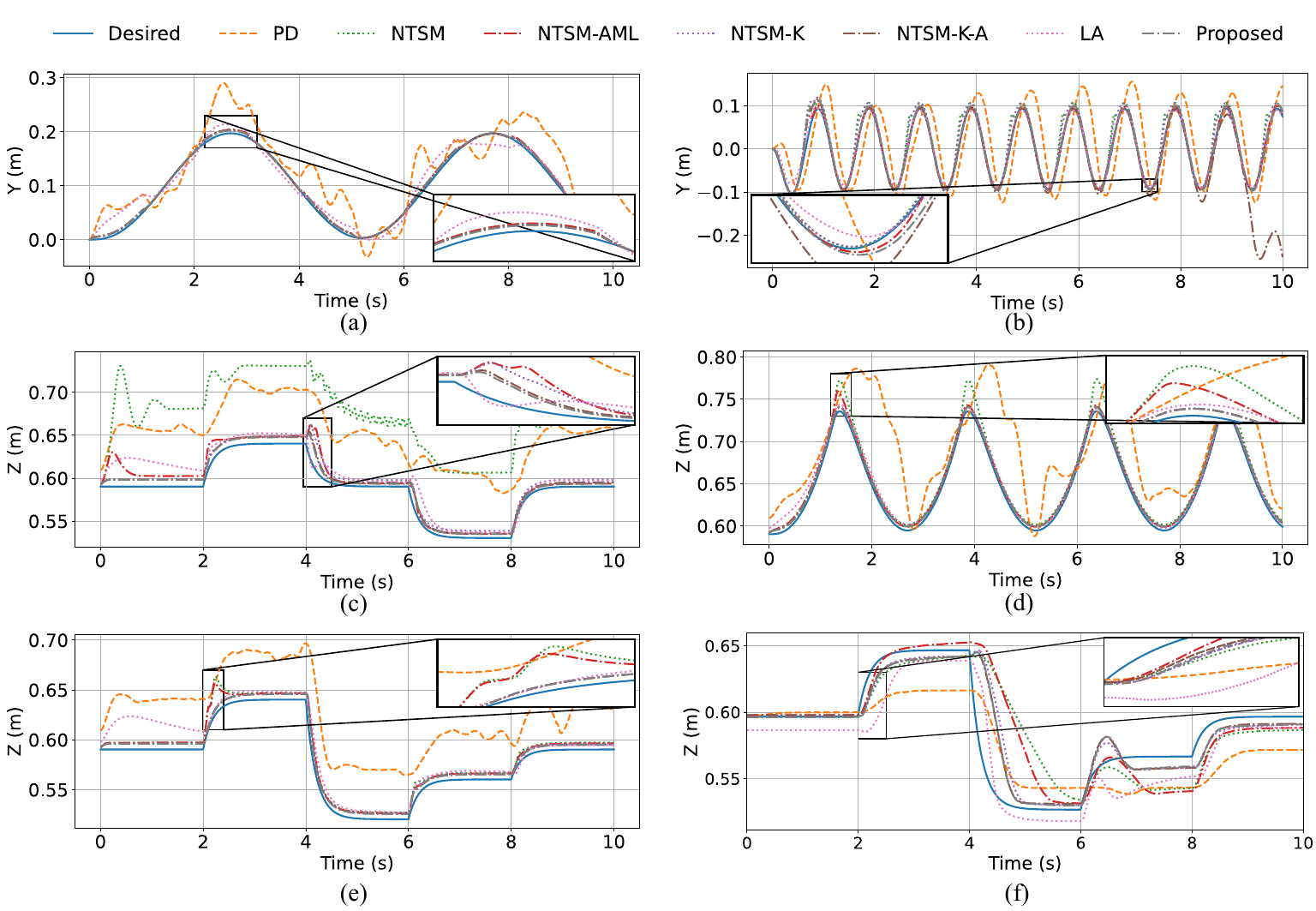}
    \caption{Tracking performance comparison of PD, LA, NTSM, NTSM-AML, NTSM-K, NTSM-K-A, and the proposed AML controller under different trajectory-force conditions. Subfigures (a)-(e) present simulation results, whereas subfigure (f) presents the real-world experimental result: (a) $T_7$, $F=18~\textrm{N}$; (b) $T_8$, $F=8~\textrm{N}$; (c) $T_9$, $F=18~\textrm{N}$; (d) $T_{10}$, $F=8~\textrm{N}$; (e) $T_{11}$, $F=7~\textrm{N}$; and (f) $T_{11}$, $F=7~\textrm{N}$ in the real-world experiment.}
    \label{fig:Controller_Comparison}
\end{figure*}
To demonstrate the advantages of the proposed AML controller in position tracking, six controllers are designed to compare with the proposed controller and exhibit ablation studies: (1). a task-space proportional-derivative (PD) controller; (2). a joint-space learning adaptive (LA) controller; (3). a joint-space NTSM controller; (4). a joint-space NTSM controller with the AML term (NTSM-AML); (5). a joint-space NTSM controller with an additional feedback term (NTSM-K); and (6). a joint-space NTSM controller with an additional feedback term and the adaptive term (NTSM-K-A). To separately evaluate the contributions of the proposed controller components, the ablation controllers are designed in a progressive manner. The comparison between NTSM and NTSM-K evaluates the contribution of the additional feedback term $K_2$, while NTSM-K and NTSM-K-A isolate the effect of the adaptive compensation. Finally, NTSM-K-A and the proposed controller evaluate the specific contribution of MTML. The six controllers are summarized as follows:

\begin{align}
    \bm{\tau}_{\textrm{PD}} = &  J(\bm{q})^{T}\Big(-K_{p_1} (\bm{p}-\bm{p}_d) - K_d J(\bm{q})\bm{\dot{q}}\Big)\nonumber\\ 
    &+ \mathcal{C}(\bm{q},\bm{\dot{q}})\bm{\dot{q}} + \mathcal{G}(\bm{q}),\\
    \bm{\tau}_{\textrm{LA}} = & \mathcal{M}(\bm q) \left(\bm{\ddot q}_{d}+\Lambda\bm{\dot e}\right) +\mathcal{C}(\bm q,\bm{\dot q})\bm{\dot q} \nonumber\\
    &+ \hat w^{M}\phi^{M}\bm{\ddot q}_{r} + \hat w^{C}\phi^{C}\bm{\dot q}_{r} + \hat w^{G}\phi^{G}\nonumber\\
    &+ K_{p_2}\bm (\bm{\dot e}+\Lambda\bm e) +K_i\int_{0}^{t}\bm{\dot e}(\xi)+\Lambda\bm e(\xi)\,d\xi\nonumber\\
    &+ K_r\textrm{tanh}\left(\frac{\bm{\dot e}+\Lambda\bm e}{\delta}\right),\\
    \bm{\tau}_{\textrm{NTSM}} = & \mathcal{M}(\bm{q}) \bm{\ddot{q}}_{d} + \mathcal{C}(\bm{q},\bm{\dot{q}})\bm{\dot{q}} + \mathcal{G}(\bm{q}) \nonumber \\
    &+ \mathcal{M}(\bm{q}) \Big( \frac{-\bm{\dot{e}}^{2-\alpha}}{\alpha \beta} - K_1 \textrm{tanh}(\kappa \bm{s}) \Big),
\end{align}
\begin{align}
    \bm{\tau}_{\textrm{NTSM-AML}} = & \mathcal{M}(\bm{q}) \bm{\ddot{q}}_{d} + \mathcal{C}(\bm{q},\bm{\dot{q}})\bm{\dot{q}} + \mathcal{G}(\bm{q} )- \Phi \bm{\hat{a}} \nonumber \\
    &+ \mathcal{M}(\bm{q}) \Big( \frac{-\bm{\dot{e}}^{2-\alpha}}{\alpha \beta} - K_1 \textrm{tanh}(\kappa \bm{s}) \Big),\\
    \bm{\tau}_{\textrm{NTSM-K}} = & \mathcal{M}(\bm{q}) \bm{\ddot{q}}_{d} + \mathcal{C}(\bm{q},\bm{\dot{q}})\bm{\dot{q}} + \mathcal{G}(\bm{q}) - K_2 \textrm{tanh}(\kappa \bm{s})\nonumber \\
    &+ \mathcal{M}(\bm{q}) \Big( \frac{-\bm{\dot{e}}^{2-\alpha}}{\alpha \beta} - K_1 \textrm{tanh}(\kappa \bm{s}) \Big),\\
    \bm{\tau}_{\textrm{NTSM-K-A}} = & \mathcal{M}(\bm{q}) \bm{\ddot{q}}_{d} + \mathcal{C}(\bm{q},\bm{\dot{q}})\bm{\dot{q}} + \mathcal{G}(\bm{q}) - K_2 \textrm{tanh}(\kappa \bm{s})\nonumber\\
    & + \mathcal{M}(\bm{q}) \Big( \frac{-\bm{\dot{e}}^{2-\alpha}}{\alpha \beta} - K_1 \textrm{tanh}(\kappa \bm{s}) \Big)\nonumber\\
    & - \bm{\hat{a}},\nonumber\\
    \dot{\hat{\bm{a}}} = & \alpha \beta \bm{\dot{e}}^{\alpha-1} \mathcal{M}(\bm{q})^{-1} \Gamma \bm{s} - \sigma \bm{\hat a}.
\end{align}
\begin{table*}[t]
    \caption{Tracking Performance Comparisons}
    \label{tab:Controller_Comparison}
    \begin{center}
    \resizebox{\textwidth}{!}{%
        \begin{tabular}{ccccccccc}
            \hline
            Force and
            & \multirow{2}{*}{Controller}
            & \multirow{2}{*}{$\mathcal{E}_x$ (mm)}
            & \multirow{2}{*}{$\mathcal{E}_y$ (mm)}
            & \multirow{2}{*}{$\mathcal{E}_z$ (mm)}
            & \multirow{2}{*}{$\mathcal{E}_{\textrm{3D}}$ (mm)}
            & \multirow{2}{*}{$\mathcal{E}_{\max}$ (mm)}
            & \multirow{2}{*}{ISU ($(\mathrm{N\,m})^2\mathrm{s}$)}
            & \multirow{2}{*}{ITAE ($\mathrm{m\,s^2}$)}
            \\
            Trajectory & & & & & & & & \\
            \hline

            \multirow{7}{*}{$F=18N,~T_{7}$}
            & PD \cite{Pollayil_2023_TRO}
            & 37.613 & 43.018 & 59.703 & 82.642 & 152.700 & 1333.656 & 3.695 \\

            & LA \cite{Evren_2019_TIMC}
            & 4.107 & 16.280 & 12.425 & 20.887 & 51.244 & 1284.929 & 0.703 \\

            & NTSM \cite{Chen_2024_TIE}
            & 38.701 & 4.067 & 84.975 & 93.461 & 178.858 & 1439.198 & 3.753 \\

            & NTSM-AML
            & 8.128 & 4.277 & 8.886 & 12.780 & 37.744 & 1268.476 & 0.430 \\

            & NTSM-K
            & 2.298 & 3.732 & 5.303 & 6.880 & 10.763
            & \textbf{1253.932} & 0.272 \\

            & NTSM-K-A
            & 2.244 & 3.683
            & \textbf{5.196}
            & 6.752
            & \textbf{10.711}
            & 1254.018
            & 0.263 \\

            \rowcolor{gray!15}
            \cellcolor{white} & Proposed
            & \textbf{2.228}
            & \textbf{3.585}
            & 5.213
            & \textbf{6.707}
            & 10.720
            & 1253.993
            & \textbf{0.263} \\
            \hline

            \multirow{7}{*}{$F=8N,~T_{8}$}
            & PD \cite{Pollayil_2023_TRO}
            & 65.206 & 61.900 & 72.722 & 115.637 & 213.745 & 1844.162 & 5.478 \\

            & LA \cite{Evren_2019_TIMC}
            & 3.259 & 6.399 & 7.810 & 10.610
            & \textbf{31.701}
            & \textbf{1350.315}
            & 0.359 \\

            & NTSM \cite{Chen_2024_TIE}
            & 39.489 & 17.974 & 70.434 & 82.725 & 115.143 & 3514.492 & 4.090 \\

            & NTSM-AML
            & 3.343
            & \textbf{5.656}
            & 5.147
            & \textbf{8.346}
            & 48.920
            & 1600.678
            & \textbf{0.317} \\

            & NTSM-K
            & 4.462 & 15.383 & 4.354 & 16.599 & 67.101 & 2203.424 & 0.601 \\

            & NTSM-K-A
            & 22.950 & 60.654 & 6.092 & 65.136 & 353.910 & 1753.905 & 1.893 \\

            \rowcolor{gray!15}
            \cellcolor{white} & Proposed
            & \textbf{3.084}
            & 8.140
            & \textbf{3.774}
            & 9.488
            & 52.606
            & 1560.674
            & 0.343 \\
            \hline

            \multirow{7}{*}{$F=18N,~T_{9}$}
            & PD \cite{Pollayil_2023_TRO}
            & 55.711 & 47.443 & 64.245 & 97.375 & 169.515
            & \textbf{1145.232}
            & 4.795 \\

            & LA \cite{Evren_2019_TIMC}
            & 4.801 & 15.873 & 15.565 & 22.744 & 48.778 & 1324.764 & 0.706 \\

            & NTSM \cite{Chen_2024_TIE}
            & 48.721 & 6.646 & 89.177 & 101.835 & 148.958 & 1664.194 & 5.026 \\

            & NTSM-AML
            & 9.425 & 6.326 & 12.275 & 16.719 & 56.267 & 1358.213 & 0.497 \\

            & NTSM-K
            & 5.661 & 5.646 & 10.235 & 12.987 & 53.989 & 1356.462 & 0.579 \\

            & NTSM-K-A
            & 4.628
            & \textbf{5.086}
            & 7.640
            & 10.279
            & 46.548
            & 1342.487
            & 0.406 \\

            \rowcolor{gray!15}
            \cellcolor{white} & Proposed
            & \textbf{4.482}
            & 5.765
            & \textbf{6.974}
            & \textbf{10.098}
            & \textbf{44.453}
            & 1343.020
            & \textbf{0.406} \\
            \hline

            \multirow{7}{*}{$F=8N,~T_{10}$}
            & PD \cite{Pollayil_2023_TRO}
            & 51.298 & 33.113 & 59.676 & 85.377 & 204.150 & 1453.686 & 3.302 \\

            & LA \cite{Evren_2019_TIMC}
            & 2.900 & 8.963 & 7.359 & 11.955 & 34.986 & 804.591 & 0.359 \\

            & NTSM \cite{Chen_2024_TIE}
            & 11.165 & 3.846 & 12.967 & 17.538 & 49.112 & 778.203 & 0.730 \\

            & NTSM-AML
            & 3.382 & 3.424 & 5.805 & 7.540 & 32.659 & 770.799 & 0.294 \\

            & NTSM-K
            & 1.786 & 3.633 & 4.656 & 6.170 & 12.171
            & \textbf{762.299}
            & 0.301 \\

            & NTSM-K-A
            & 1.809
            & \textbf{3.290}
            & 4.718
            & 6.030
            & \textbf{12.113}
            & 766.149
            & 0.294 \\

            \rowcolor{gray!15}
            \cellcolor{white} & Proposed
            & \textbf{1.668}
            & \textbf{3.290}
            & \textbf{4.529}
            & \textbf{5.841}
            & 13.187
            & 763.609
            & \textbf{0.278} \\
            \hline

            \multirow{7}{*}{$F=7N,~T_{11}$}
            & PD \cite{Pollayil_2023_TRO}
            & 14.611 & 14.080 & 47.587 & 51.732 & 107.411 & 584.538 & 2.541 \\

            & LA \cite{Evren_2019_TIMC}
            & 3.449 & 2.434 & 14.159 & 14.775 & 34.152 & 126.399 & 0.426 \\

            & NTSM \cite{Chen_2024_TIE}
            & 5.762 & 1.400 & 9.131 & 10.888 & 47.598
            & \textbf{98.033}
            & 0.442 \\

            & NTSM-AML
            & 4.550 & 1.394 & 8.144 & 9.432 & 41.843 & 98.068 & 0.365 \\

            & NTSM-K
            & 1.537 & 1.198 & 6.323 & 6.617 & 26.096 & 102.225 & 0.317 \\

            & NTSM-K-A
            & \textbf{1.254}
            & 1.184
            & 5.851
            & 6.100
            & 25.928
            & 102.218
            & 0.277 \\

            \rowcolor{gray!15}
            \cellcolor{white} & Proposed
            & 1.256
            & \textbf{1.181}
            & \textbf{5.847}
            & \textbf{6.095}
            & \textbf{25.926}
            & 102.128
            & \textbf{0.277} \\
            \hline

            \multirow{7}{*}{\shortstack{$F=7N,~T_{11}$\\(Experiment)}}
            & PD \cite{Pollayil_2023_TRO}
            & 21.789 & 19.666 & 24.200 & 38.041 & 76.794 & 3928.065 & 1.915 \\

            & LA \cite{Evren_2019_TIMC}
            & 13.182 & 12.485 & 17.270 & 25.058
            & \textbf{69.359}
            & 4048.799
            & 1.143 \\

            & NTSM \cite{Chen_2024_TIE}
            & 25.992 & 12.654 & 27.787 & 40.098 & 98.369
            & \textbf{3764.495}
            & 1.637 \\

            & NTSM-AML
            & 25.041 & 12.076 & 26.144 & 38.163 & 99.197 & 3797.832 & 1.558 \\

            & NTSM-K
            & 11.459
            & \textbf{6.163}
            & 18.276
            & 22.435
            & 87.294
            & 4005.619
            & 0.731 \\

            & NTSM-K-A
            & \textbf{10.531}
            & 6.766
            & 16.408
            & \textbf{20.637}
            & 79.539
            & 3899.129
            & 0.686 \\

            \rowcolor{gray!15}
            \cellcolor{white} & Proposed
            & \textbf{10.531}
            & 6.982
            & \textbf{16.379}
            & 20.686
            & 80.351
            & 3952.836
            & \textbf{0.682} \\
            \hline

        \end{tabular}%
    }
    \end{center}
\end{table*}

For the PD controller, the proportional and derivative gains are set to the high stiffness in~\cite{Pollayil_2023_TRO} as $K_{p_1}=390$ and $K_d=39.5$, respectively. For the learning adaptive controller in ~\cite{Evren_2019_TIMC}, the gains are set to $K_{p_2}=20$, $K_i=10$, and $K_r=1$. For the NTSM-based controllers, the common parameters are selected as $\alpha=1.66$, $\beta=0.2$, $\kappa=20$, and $K_1=40$ to ensure a fair comparison. The NTSM-K and the proposed AML controller further employ $K_2=\{1,1,1,1,5,5,5\}$. For the adaptive controllers NTSM-AML, NTSM-K-A, and the proposed AML controller, the adaptive gain is chosen as $\Gamma=0.1I_7$ with $\sigma=1\times10^{-5}$. The inverse kinematics solver used in this paper is adopted from~\cite{He_2021_ICCMA}, with Joint 7 fixed to prevent infinite solutions.
\begin{figure*}[t]
    \centering
    \subfloat[]{
        \includegraphics[width=0.5\linewidth]{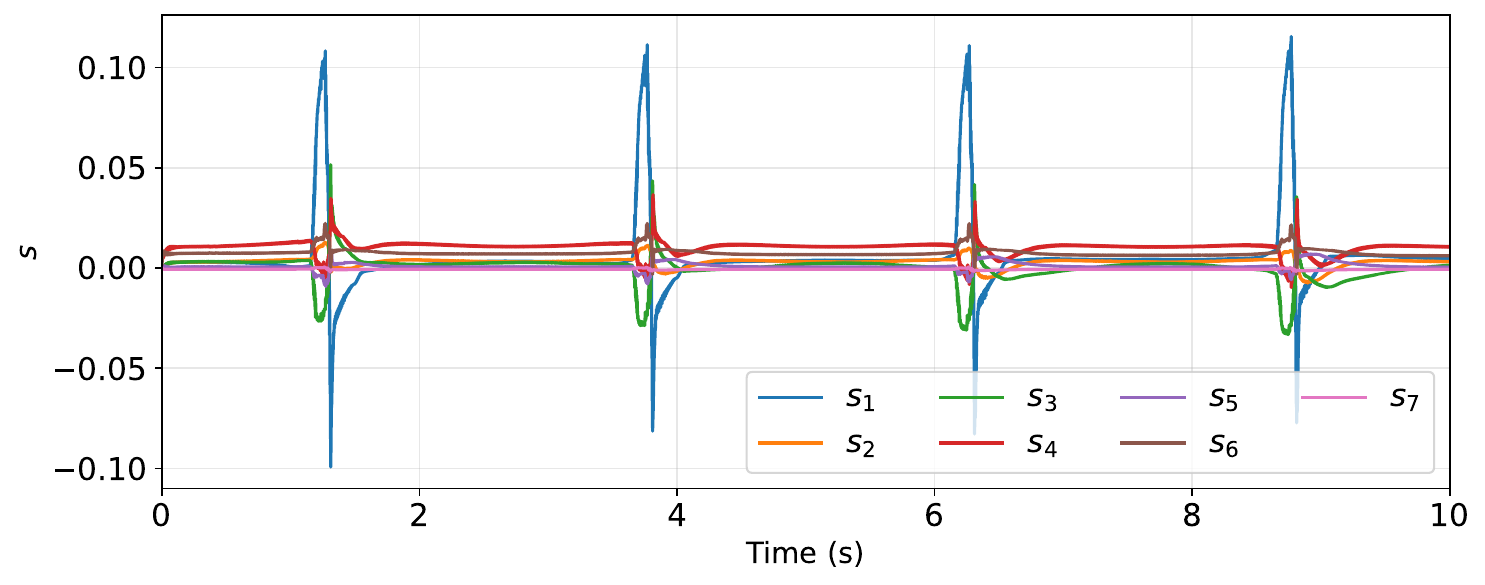}}
    \subfloat[]{
        \includegraphics[width=0.5\linewidth]{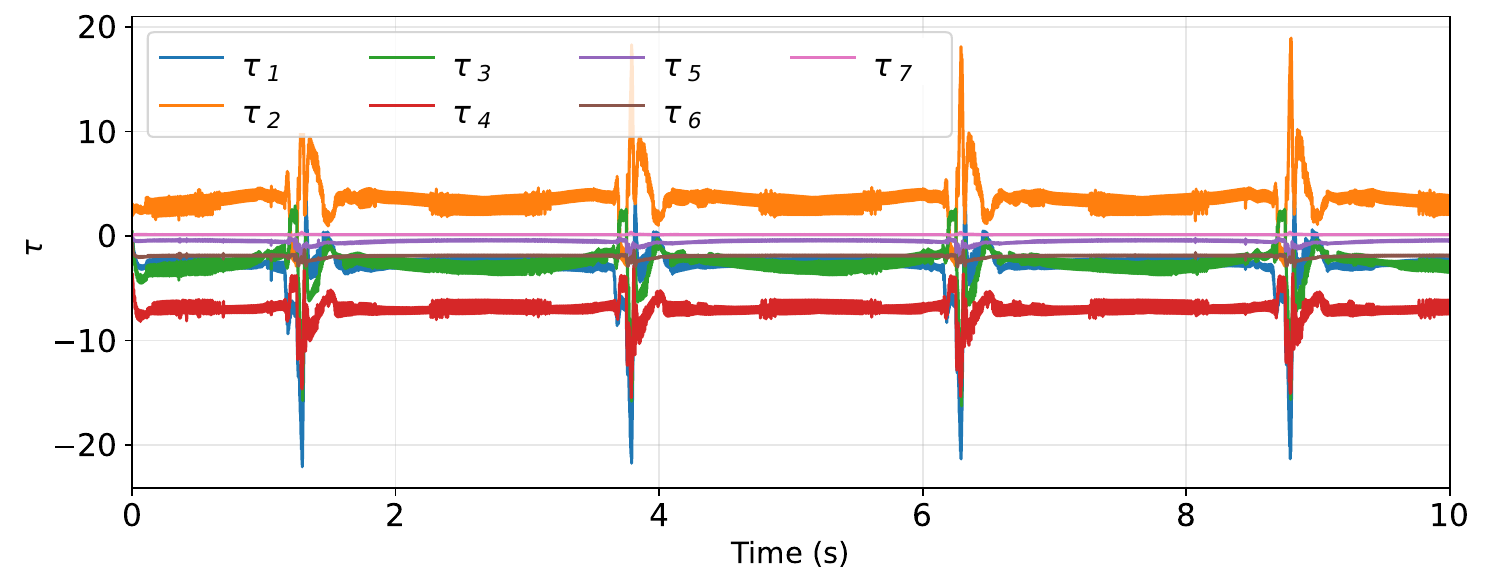}}\\
    \subfloat[]{
        \includegraphics[width=0.5\linewidth]{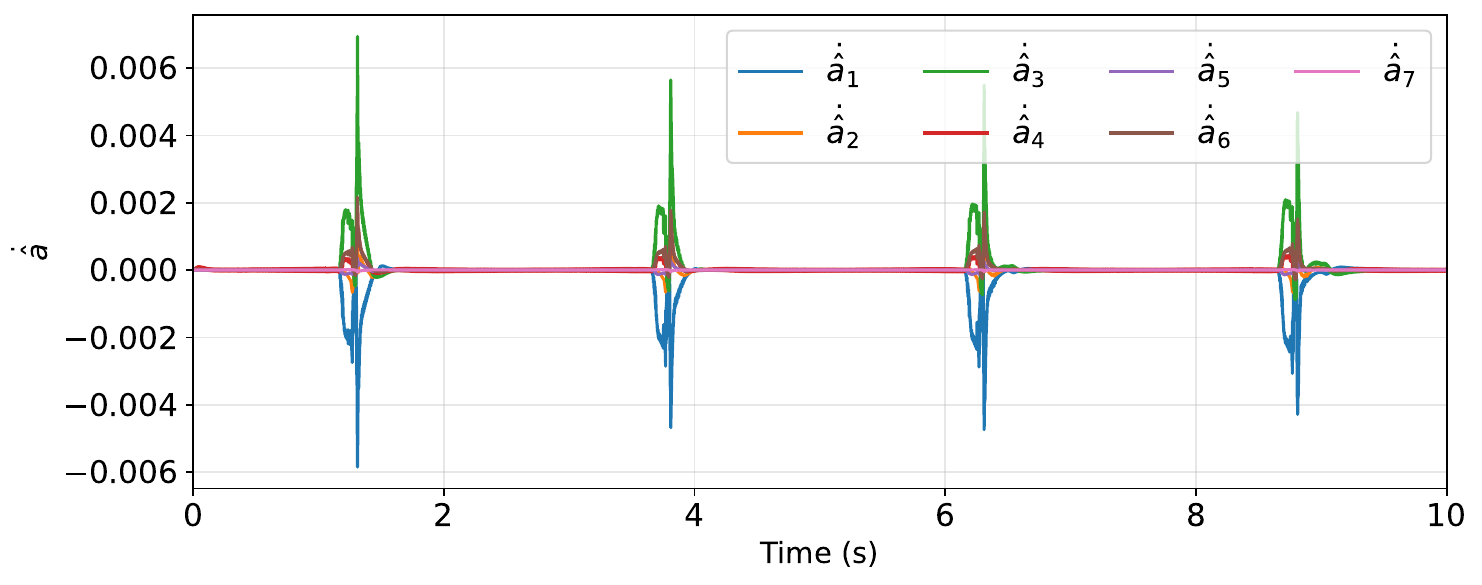}}
    \subfloat[]{
        \includegraphics[width=0.5\linewidth]{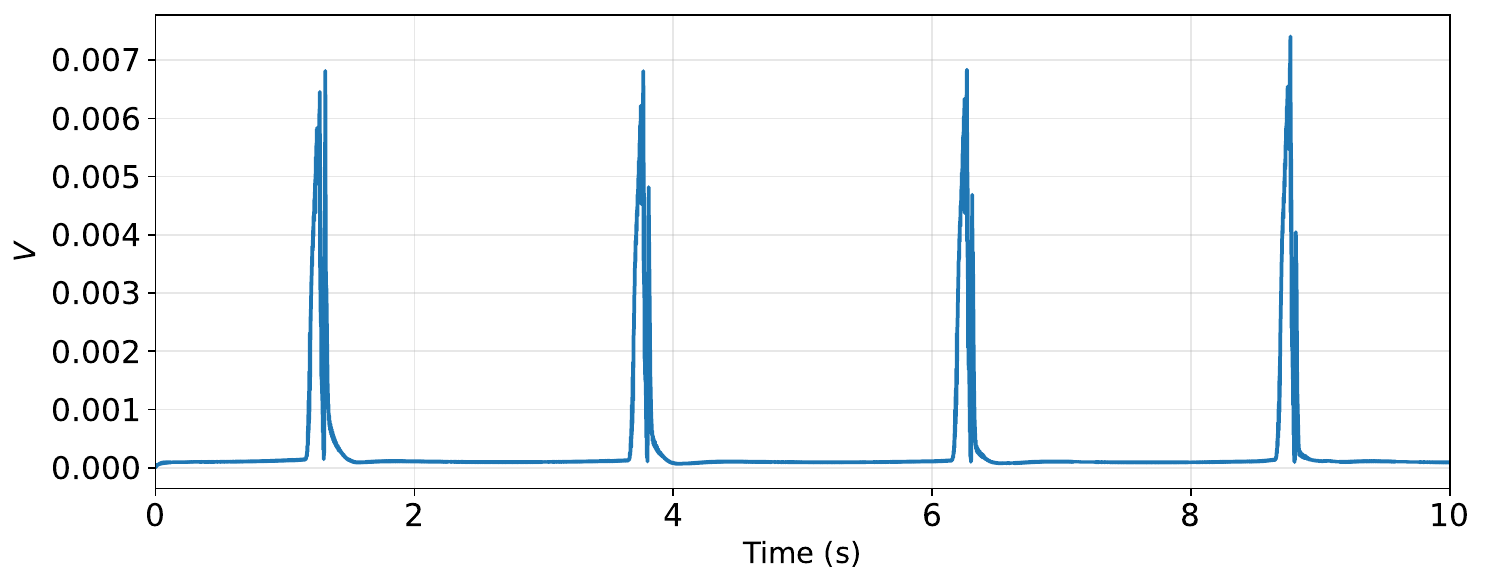}}\\
    \caption{Convergence-related variables of the proposed controller under $F=8~\mathrm{N}$ and trajectory $T_{10}$: (a) sliding surface $\bm{s}$; (b) control torque $\bm{\tau}$; (c) adaptive parameter update rate $\bm{\dot{\hat{a}}}$; and (d) Lyapunov function $V_s$.}
    \label{fig:Convergence_Analysis}
\end{figure*}
\begin{figure*}[t]
    \centering
    \subfloat[]{
        \includegraphics[width=0.68\linewidth]{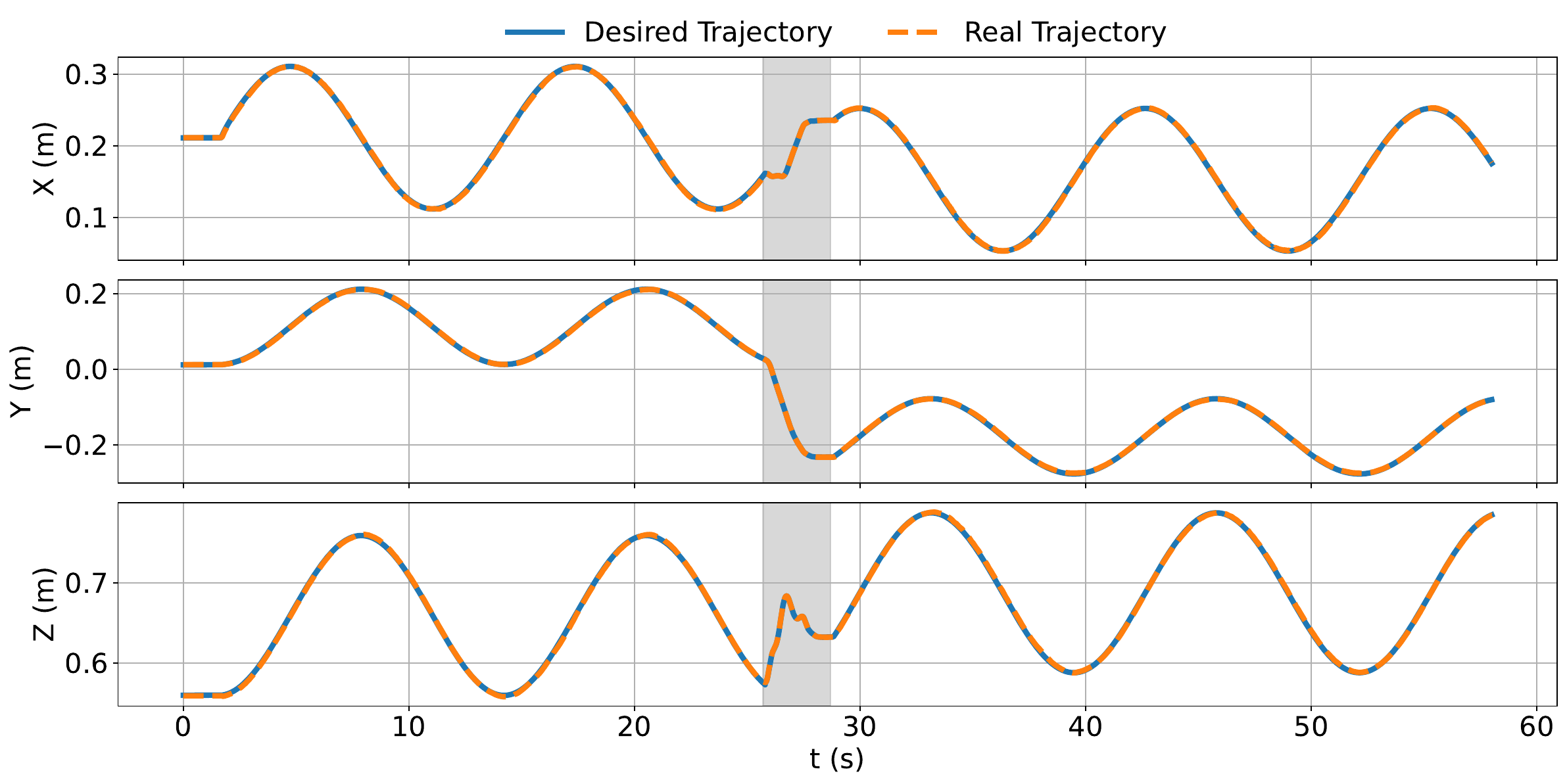}}
    \subfloat[]{
        \includegraphics[width=0.32\linewidth]{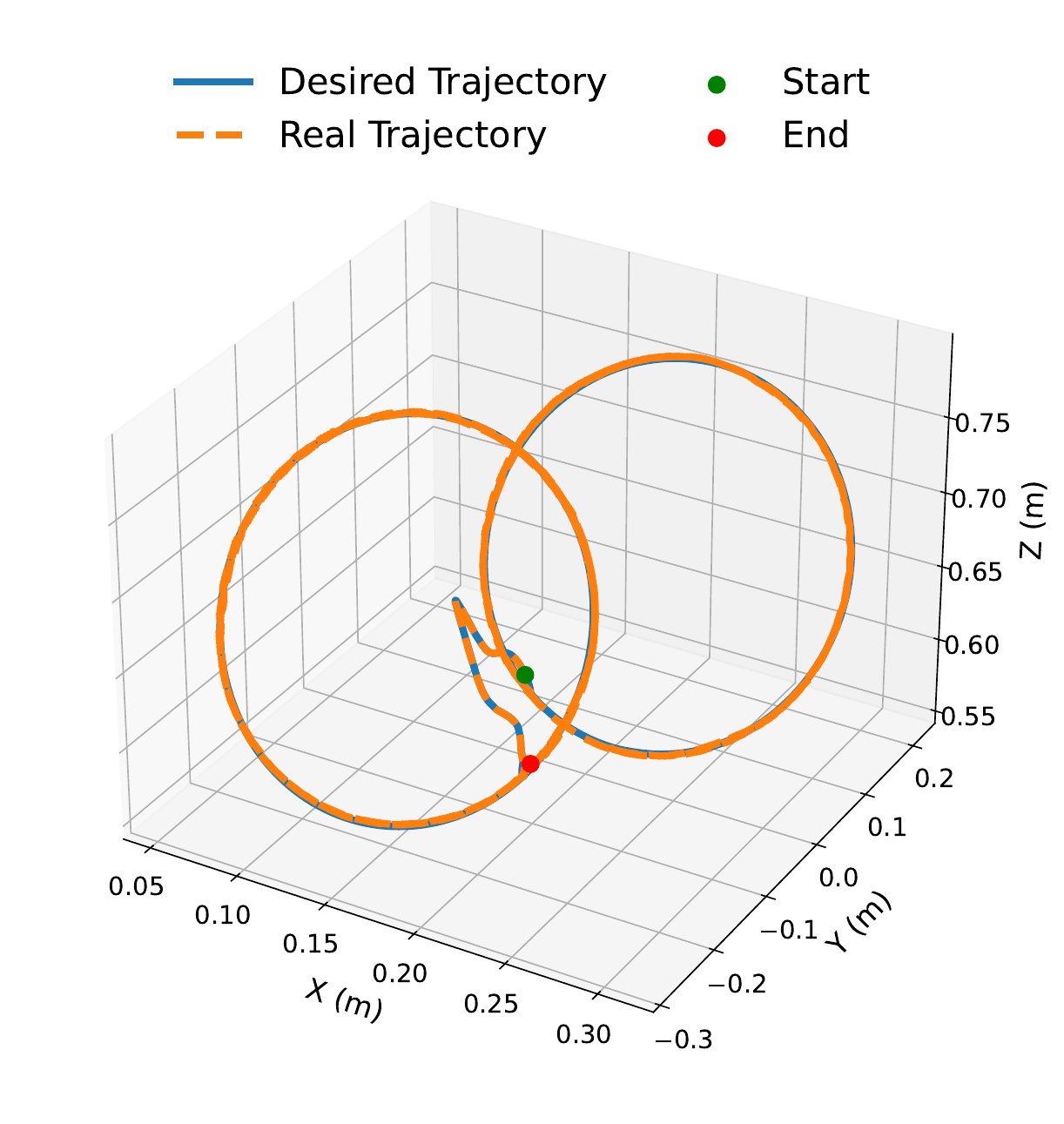}}
    \caption{Integrated framework validation during physical HRI with transitions between active and passive interaction modes. (a) Trajectories along the x-, y-, and z-axes, in which the gray shaded area indicates the active mode; (b) Trajectory in 3D space.}
    \label{fig:Real_Trajectory}
\end{figure*}

The comparative results of the PD, LA, NTSM, NTSM-AML, NTSM-K, NTSM-K-A, and proposed controllers are presented in Fig.~\ref{fig:Controller_Comparison} and Table~\ref{tab:Controller_Comparison}. Six tracking tasks are designed for comparison, including five simulation tasks and one real-world experiment. The simulation tasks are configured as follows: Task 1 employs the trajectory $T_7$ with an external force of $F=18~\textrm{N}$ whose direction switches every two seconds; Task 2 employs trajectory $T_8$ with a constant three-dimensional (3D) force of $F=8~\textrm{N}$; Task 3 employs trajectory $T_9$ with a constant 3D force of $F=18~\textrm{N}$; Task 4 employs trajectory $T_{10}$ with a constant 3D force of $F=8~\textrm{N}$; and Task 5 employs trajectory $T_{11}$ with a constant $z$-axis force of $F=7~\textrm{N}$. Task 6 is conducted on the real Franka robot using the same trajectory $T_{11}$ and constant z-axis force of $F=7~\textrm{N}$ as in Task 5, thereby enabling a direct comparison between the simulations and real-world performance. All trajectories and force conditions differ from those used during training to evaluate the generalization capability of the proposed controller. The six tasks impose different performance requirements on the controller. In Task 1, the force direction switches every two seconds, requiring rapid adaptation to varying external disturbances, whereas Task 2 requires faster robot motion. Tasks 3, 5, and 6 evaluate tracking performance under non-periodic trajectories, while Task 4 evaluates tracking performance under varying acceleration conditions.

The tracking performance is evaluated using four metrics. For each Cartesian direction $j\in\{x,y,z\}$, the root mean square error (RMSE) is defined as
\begin{equation}
    \mathcal{E}_{j}
    =
    \sqrt{
    \frac{1}{N}
    \sum_{i=1}^{N}
    \left(p_{j}(i)-p_{d,j}(i)\right)^2
    },
\end{equation}
where $N$ denotes the number of sampled data points. The overall 3D tracking error is further evaluated as
\begin{equation}
    \mathcal{E}_{\textrm{3D}}
    =
    \sqrt{
    \frac{1}{N}
    \sum_{i=1}^{N}
    \|\bm{p}(i)-\bm{p}_{d}(i)\|_2^2
    },
\end{equation}
while the maximum position error is defined as
\begin{equation}
    \mathcal{E}_{\max}
    =
    \max_{i}
    \|\bm{p}(i)-\bm{p}_{d}(i)\|_2 .
\end{equation}

In addition to the tracking accuracy, the integral of squared control input (ISU) is employed to quantify the required control effort:
\begin{equation}
    \textrm{ISU}
    =
    \int_{0}^{T}
    \bm{\tau}^{T}(t)\bm{\tau}(t)\,dt,
\end{equation}
and the integral of time-weighted absolute error (ITAE) is used to penalize tracking errors that persist for a longer duration:
\begin{equation}
    \textrm{ITAE}
    =
    \int_{0}^{T}
    t\|\bm{p}(t)-\bm{p}_{d}(t)\|_2\,dt.
\end{equation}

Therefore, $\mathcal{E}_{x}$, $\mathcal{E}_{y}$, $\mathcal{E}_{z}$, and $\mathcal{E}_{\textrm{3D}}$ characterize the overall tracking accuracy, $\mathcal{E}_{\max}$ reflects the worst-case deviation, ISU evaluates the control effort efficiency, and ITAE emphasizes both the magnitude and duration of the tracking error.

Overall, the proposed controller achieves consistently low tracking errors under the six unseen trajectory-force conditions, with an average $\mathcal{E}_{\textrm{3D}}$ of $9.819$~mm. Relative to the conventional NTSM controller, the proposed method reduces the average $\mathcal{E}_{\textrm{3D}}$ and ITAE by $83.0\%$ and $85.7\%$, respectively. The effectiveness of the AML component is further demonstrated by comparing NTSM-AML with NTSM: introducing AML reduces the average $\mathcal{E}_{\textrm{3D}}$ from $57.757$~mm to $15.497$~mm and the average ITAE from $2.613$ to $0.577$, corresponding to reductions of $73.2\%$ and $77.9\%$, respectively. The ablation results with NTSM-K and NTSM-K-A further indicate that the complete proposed framework provides more consistent tracking performance across different motion and disturbance conditions. Compared with NTSM-K, the proposed controller reduces the average $\mathcal{E}_{\textrm{3D}}$, $\mathcal{E}_{\max}$, and ITAE by $17.8\%$, $11.7\%$, and $19.7\%$, respectively, while maintaining a comparable overall control effort.

To further illustrate the convergence behavior of the proposed controller, the key variables involved in the stability analysis are presented in Fig.~\ref{fig:Convergence_Analysis} for the case of $F=8~\mathrm{N}$ and trajectory $T_{10}$. As shown in Fig.~\ref{fig:Convergence_Analysis}(a), the sliding surfaces remain bounded and converge to zero after each transient response. Fig.~\ref{fig:Convergence_Analysis}(b) shows that control torque $\bm{\tau}$ remains bounded throughout the trajectory despite the short peaks. Correspondingly, the adaptive parameter update rates $\bm{\dot{\hat{a}}}$ exhibit transient adjustments and rapidly
converge to zero, as shown in Fig.~\ref{fig:Convergence_Analysis}(c). Moreover, the Lyapunov function $V_s$ remains positive and returns to a small value following each transient, as illustrated in Fig.~\ref{fig:Convergence_Analysis}(d). These results are consistent with the bounded and convergent behavior predicted by the convergence analysis.

The closed loop HRI mode switching experiment results are shown in Fig.\ref{fig:Real_Trajectory}. The mode recognition NN $\mathcal{N}$ rapidly detects human intention when the human is actively dragging the robot, while the robust AML controller effectively reduces the errors caused by human interaction and environmental disturbances when the human operates in the passive mode. An example of the demonstrations can be seen in {https://youtu.be/yGcBRhp8WIA}.

\section{Conclusions}
A novel framework integrating mode recognition, environment-basis NN training, and adaptive robot control has been developed for HRI. The proposed mode recognition NN eliminates the need for manually defined thresholds. By coupling the adaptive controller with the mode recognition, the adaptively estimated force-dependent parameters are used as complementary interaction information, improving the recognition rates from $90.1\%$ to $99.1\%$ for the passive mode and from $79.8\%$ to $97.1\%$ for the active mode. A meta-learning algorithm is developed to train the environment-basis NN using multiple trajectories, ensuring that the parameters computed by the network are force-dependent only. The proposed adaptive meta-learning controller achieves an average 3D RMSE of $9.819$~mm over six unseen trajectory-force conditions and the ablation studies demonstrate the benefit of the proposed controller components.

The proposed framework is validated on one interaction task, and its generalization to more complex tasks and different robots requires further investigation. In addition, the performance may depend on the quality and diversity of the training data used in the meta-learning process. In future work, the effects of measurement noise will be investigated to further evaluate the robustness of the proposed framework. More complex manipulation tasks with changing payloads and uncertain contacts would provide a stronger demonstration of the practical applicability.


\bibliographystyle{Bibliography/IEEEtranTIE}
\bibliography{Bibliography/IEEEabrv,Bibliography/BIB_25-TMech-xxxx}\ 

\vspace{-1cm}
\begin{IEEEbiography}
[{\includegraphics[width=1in,height=1.25in,clip,keepaspectratio]{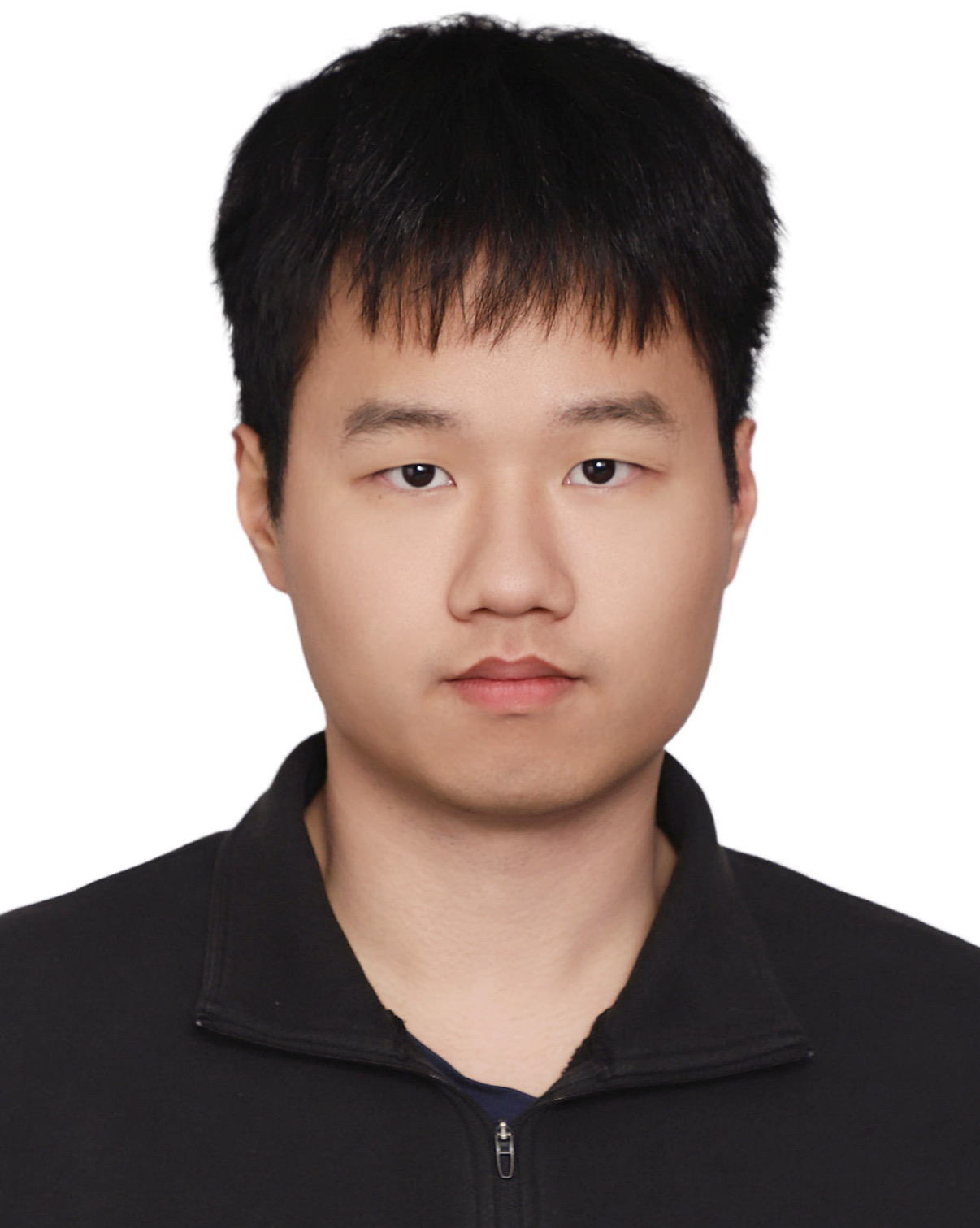}}]{Nuo Chen} (Student Member, IEEE) received the B.E. degree in robotics engineering from Southern University of Science and Technology, Shenzhen, China, in 2022, and M.A.Sc. degree in mechanical engineering from Dalhousie University, Halifax, NS, Canada, in 2024. He is currently working toward the Ph.D. degree in mechanical engineering with Dalhousie University, Halifax, NS, Canada.

His research interests include robotics control and machine learning.
\end{IEEEbiography}
\vspace{-1cm}
\begin{IEEEbiography}[{\includegraphics[width=1in,height=1.25in,clip,keepaspectratio]{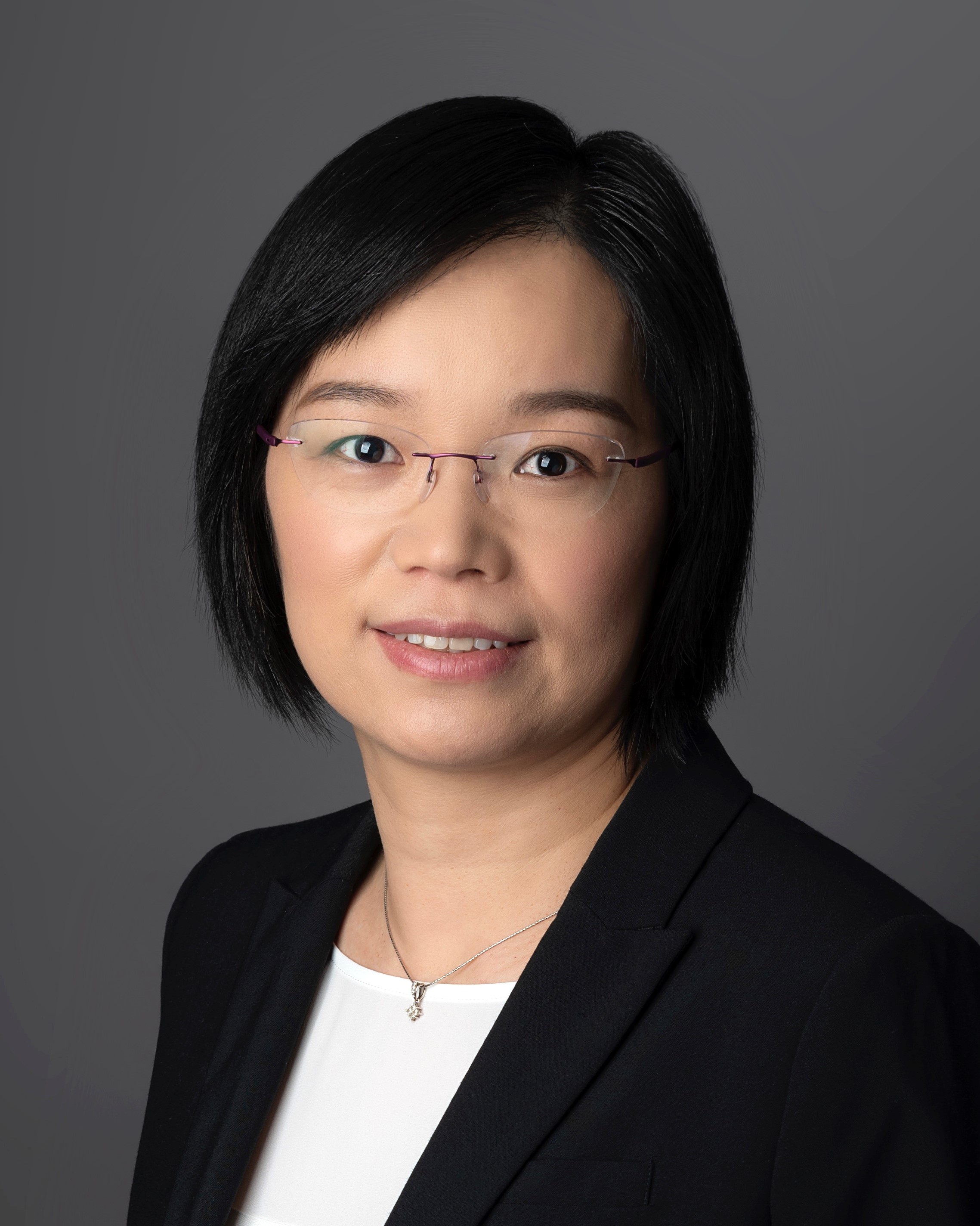}}]{Ya-Jun Pan} (Senior Member, IEEE) received the B.E. degree in mechanical engineering from Yanshan University, Qinhuangdao, China, in 1996, the M.E. degree in mechanical engineering from Zhejiang University, Hangzhou, China, in 1999, and the Ph.D. degree in electrical and computer engineering from the National University of Singapore, Singapore, in 2003. She is a Professor with the Department of Mechanical Engineering, Dalhousie University, Halifax, NS, Canada. She held the post-doctoral fellow positions with the Laboratoire d'Automatique de Grenoble, CNRS, Grenoble, France and the Department of Electrical and Computer Engineering, University of Alberta, Edmonton, Canada, respectively. Her research interests include robust nonlinear control, cyber-physical systems, intelligent transportation systems, haptics, and collaborative multiple robotic systems. She is a Fellow of the Canadian Academy of Engineering (CAE), a Fellow of the Engineering Institute of Canada (EIC), a Fellow of ASME, a Fellow of CSME, a Senior Member of IEEE, and a registered Professional Engineer in Nova Scotia, Canada.
\end{IEEEbiography}

\end{document}